\documentclass[10pt,twocolumn,letterpaper]{article}
\usepackage[pagenumbers]{wacv} 
\usepackage{graphicx}
\usepackage{amsmath}
\usepackage{amssymb}
\usepackage{booktabs}
\usepackage{comment}
\usepackage{multirow}
\usepackage{bm}
\usepackage{colortbl}
\usepackage{caption}
\usepackage{enumitem} 
\usepackage{lipsum}
\usepackage{wrapfig}

\usepackage[utf8]{inputenc} 
\usepackage[T1]{fontenc}    

\usepackage{url}        
\usepackage{booktabs}       
\usepackage{amsfonts}    
\usepackage{nicefrac}     
\usepackage{microtype}
\usepackage{xcolor}

\DeclareMathOperator*{\argmax}{arg\,max}

\usepackage{xargs}

\definecolor{red}{rgb}{1, 0.7, 0.7} 
\definecolor{orange}{rgb}{1, 0.85, 0.7} 
\definecolor{yellow}{rgb}{1, 1, 0.7} 
\definecolor{orange(webcolor)}{rgb}{1.0, 0.65, 0.0}
\definecolor{darkblue}{rgb}{0.0, 0.0, 0.55}

\definecolor{tabfirst}{rgb}{1, 0.7, 0.7} 
\definecolor{tabsecond}{rgb}{1, 0.85, 0.7} 
\definecolor{tabthird}{rgb}{1, 1, 0.7}

\usepackage[pagebackref,breaklinks,colorlinks]{hyperref}

\usepackage[capitalize]{cleveref}
\crefname{section}{Sec.}{Secs.}
\Crefname{section}{Section}{Sections}
\Crefname{table}{Table}{Tables}
\crefname{table}{Tab.}{Tabs.}

\def\wacvPaperID{928} 
\def\confName{WACV}
\def\confYear{2025}

\begin{document}
\title{Transientangelo: Few-Viewpoint Surface Reconstruction\\Using Single-Photon Lidar}

\author{%
  Weihan Luo$^{1}$ \\
  {weihan.luo@mail.utoronto.ca}
  \and
  Anagh Malik$^{1,2}$ \\
  {anagh@cs.toronto.edu}
  \and
David B. Lindell$^{1,2}$ \\
  {lindell@cs.toronto.edu}
\and
  {\normalfont $^1$University of Toronto \hspace{10pt}}
  {\normalfont $^2$Vector Institute \hspace{10pt}} \\
\url{https://weihan1.github.io/transientangelo/}
}

\twocolumn[{%
\renewcommand\twocolumn[1][]{#1}%
\maketitle
\begin{center}
    \captionsetup{type=figure}
    \vspace{-0.7cm}
        \includegraphics[width=\textwidth]{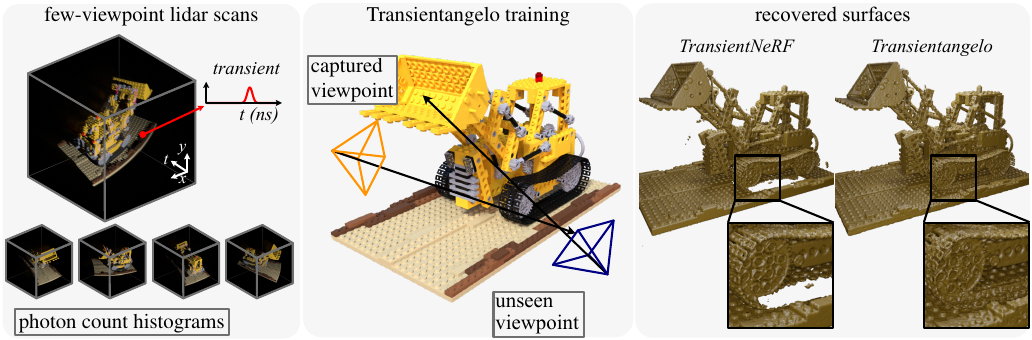}
    \\
    \captionof{figure}{
Transientangelo takes as input raw lidar scans from sparse viewpoints. These scans are used to optimize a scene representation based on a signed distance function, which is further regularized to constrain the geometry from both captured and unseen viewpoints. The method recovers higher-fidelity surfaces than previous methods in the sparse-view and low-photon regime (i.e., from tens to hundreds of measured photons per pixel). The above scene was trained using simulated single-photon lidar data from five viewpoints, with an average of 150 photons per pixel over the occupied regions of the scene.}
    \label{fig:teaser}
\end{center}
}]

\maketitle

\begin{abstract}
We consider the problem of few-viewpoint 3D surface reconstruction using raw measurements from a lidar system. Lidar captures 3D scene geometry by emitting pulses of light to a target and recording the speed-of-light time delay of the reflected light. 
However, conventional lidar systems do not output the raw, captured waveforms of backscattered light; instead, they preprocess these data into a 3D point cloud.
Since this procedure typically does not accurately model the noise statistics of the system, exploit spatial priors, or incorporate information about downstream tasks, it ultimately discards useful information that is encoded in raw measurements of backscattered light. Here, we propose to leverage raw measurements captured with a single-photon lidar system from multiple viewpoints to optimize a neural surface representation of a scene.
The measurements consist of time-resolved photon count histograms, or \textit{transients}, which capture information about backscattered light at picosecond time scales. Additionally, we develop new regularization strategies that improve robustness to photon noise, enabling accurate surface reconstruction with as few as 10 photons per pixel. Our method outperforms other techniques for few-viewpoint 3D reconstruction based on depth maps, point clouds, or conventional lidar as demonstrated in simulation and with captured data.
\end{abstract}

\section{Introduction}
\label{sec:intro}

Within the last few years, significant progress has been made on the problem of multi-view 3D surface reconstruction.
Recent methods combine efficient neural surface representations with volumetric rendering models, enabling high-fidelity scene reconstruction from multi-view images~\cite{li2023neuralangelo, tewari2022advances}.
Overall, achieving accurate scene reconstruction is now relatively straightforward when tens to hundreds of input images are available. 
However, a significant remaining challenge is achieving robust reconstruction when the number of input viewpoints is reduced to only a few (e.g., less than ten).  
In this case, scene reconstruction from conventional images alone is ill-posed~\cite{kutulakos2000space} and requires strong priors to solve robustly~\cite{wu2023reconfusion}.

Instead of using conventional images, we investigate this challenge using lidar measurements captured from a few (2--5) different viewpoints. Lidar measures the distance to each scene point by illuminating it with a pulse of light and measuring the speed-of-light time delay of the back-reflected pulse.
These measurements provide additional constraints to the multi-view reconstruction problem. However, conventional lidar systems do not output the raw, captured waveforms of backscattered light; instead, they post-process these data into a 3D point cloud. Since this procedure typically does not accurately model the noise statistics of the system, exploit spatial priors, or incorporate information about downstream tasks, it ultimately discards useful information that is encoded in raw measurements of backscattered light. We are interested in using the raw measurements captured by a single-photon lidar system---the time-resolved photon counts histograms, or \textit{transients}, which contain information about the precise arrival times of backscattered light particles at picosecond time scales. 

Previous methods for multi-viewpoint 3D reconstruction---whether based on conventional images or lidar---have several drawbacks.
First, image-based methods are usually trained with tens or hundreds of images, making acquisition difficult~\cite{muller2022instant,mildenhall2021nerf,wang2021neus,li2023neuralangelo}.
While several recent works attempt to tackle the few-view reconstruction problem~\cite{niemeyer2022regnerf,malik2023transient,wang2023sparsenerf,yu2022monosdf}, they typically fail for the case where the baseline between input images becomes large, and the problem becomes increasingly ill-posed. 
Recent methods that leverage large-scale diffusion models~\cite{wu2023reconfusion, gao2024cat3d} alleviate this limitation to some extent, but are also prone to hallucinating scene geometry.
Other techniques improve few-view 3D reconstruction by introducing additional geometric constraints from point clouds~\cite{rematas2022urban} or monocular depth estimators~\cite{yu2022monosdf}.
In this vein, our work is closest to TransientNeRF~\cite{malik2023transient}, which shows that using raw lidar data instead of depth maps or point clouds leads to significant improvements in reconstructed scene appearance in the few-view setting. 
However, TransientNeRF is based on a volumetric representation and is designed for appearance modeling rather than surface reconstruction. Further, TransientNeRF uses long acquisition times ($\approx$20 minutes per view) to capture per-pixel photon count histograms with several thousand photon arrivals each.

To address these limitations we propose Transientangelo, a method for few-viewpoint scene reconstruction using transient supervision and a scene representation based on signed distance functions (SDFs). 
Inspired by Neuralangelo~\cite{li2023neuralangelo}, we use an efficient neural representation with a hashing-based feature grid to parameterize an SDF and radiance field. 
The representation is then trained using supervision from photon count histograms.
To further constrain the reconstructed surfaces, we regularize the scene geometry from unseen viewpoints and use supervision from the captured reflectivity (i.e., the time-integrated lidar measurements).

Overall, we improve few-viewpoint 3D surface reconstruction from raw lidar measurements compared to previous work (see Figure~\ref{fig:teaser}), while also demonstrating significant improvements in reconstructed geometry compared to image-based or depth-based approaches. Our approach also improves robustness to lidar measurements captured with relatively few photon counts (e.g., 10–300), which is especially important when operating at fast acquisition speeds or when imaging targets at long range.
In summary, we make the following technical contributions.
\begin{itemize}
    \item We introduce Transientangelo: a method to perform surface reconstruction using few-viewpoint single-photon lidar. 
    \item We propose regularization techniques that improve the reconstructed geometry and provide improved robustness in the low-flux regime where only tens to hundreds of photons are collected per pixel.

    \item We introduce a multi-viewpoint simulated and captured transient dataset with a range of photon count levels, which we use to demonstrate state-of-the-art performance for few-viewpoint surface reconstruction.
\end{itemize}

\section{Related Work}
\label{gen_inst}
\begin{figure*}[th!]
    \begin{center}
    \includegraphics[width=\textwidth]{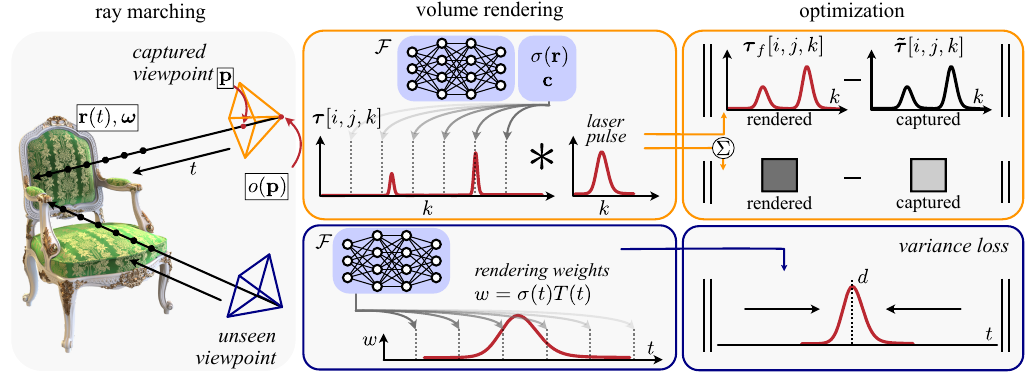}
    \end{center}
    \caption{Transientangelo training procedure. For a pixel $\mathbf{p}$, we cast out a ray $\mathbf{r}(t) = \mathbf{o}+tc\bm{\omega}(\mathbf{p})$. \textcolor{orange(webcolor)}{\textbf{Captured viewpoint:}} 3D ray coordinates are passed through the neural surface representation $\mathcal{F}$ to retrieve a radiance and an SDF value, which gets converted to density $\sigma$ (see Equation ~\ref{eq:density}). These values are then binned into a transient, which, after a convolution with the laser pulse, gives the final rendered transient $\boldsymbol{\tau}_f[i, j, k]$. The network is supervised with an L1 loss between the rendered and captured transient and an L1 loss between the integrals of the rendered and captured transients (reflectivity loss). \textcolor{darkblue}{\textbf{Unseen viewpoint:}} 3D ray coordinates are passed through the neural representation to retrieve rendering weights, which are used to calculate the variance of the weights around the depth $d$. The network is trained to minimize this variance, resulting in thinner surfaces and removing spurious zero-level sets.}
    \vspace{-10px}
    \label{fig:method}
\end{figure*}

Transientangelo brings together the areas of active single-photon imaging, neural surface reconstruction, and few-viewpoint 3D reconstruction. 

\paragraph{Active single-photon imaging.}
We are specifically concerned with lidar systems based on single-photon avalanche diodes (SPADs), which output precise timestamps corresponding to the arrival times of individual captured photons. 
For SPAD-based lidar, a transient is measured by repeatedly illuminating a point with pulses of light and accumulating the individual photon arrival times into a time-resolved histogram. 
Such captured histograms of a scene can be used for direct scene reconstruction~\cite{lindell2018towards,jungerman20223d,klinghoffer2023platonerf,malik2023transient}, non-line of sight scene reconstruction~\cite{chen2019learning,otoole2018confocal,rapp2020seeing,xin2019theory,faccio2020non}, seeing through participating media~\cite{lindell2020three}, visualizations of light propagation~\cite{malik2024flying,velten2013femto,gariepy2015single} or to uncover statistical properties of captured photons~\cite{rapp2017few,rapp2021high}.
Our work differs from these in that we specifically aim to recover high-fidelity surfaces from multi-viewpoint SPAD data.
We are motivated by previous work\cite{malik2023transient} that demonstrates the strong geometric information encoded in transients and their promise for multi-viewpoint surface reconstruction. 

While many previous techniques on 3D imaging with SPADs use high-powered lasers or long acquisition times to capture hundreds or thousands of photons~\cite{malik2023transient, lindell2018towards, lindell2020three,otoole2018confocal}, our work is more closely related to single-photon imaging techniques that operate in the photon-starved regime. 
Such methods exploit spatial or temporal correlations in the photon arrivals to recover depth or reflectivity from only a few captured photons at each pixel~\cite{lee2023caspi,rapp2017few}.
Instead, our approach exploits cross-viewpoint information and reflectivity-based regularization for scene reconstruction from as few as 10 collected photons per pixel.

\paragraph{Neural surface reconstruction.}

Neural surface representations typically use either occupancy~\cite{oechsle2021unisurf, niemeyer2020differentiable, peng2020convolutional} or a signed distance function (SDF)~\cite{wang2021neus, wang2023neus2, li2023neuralangelo, yariv2021volume, yariv2020multiview} to model scene geometry.
The former uses a neural network to classify points in 3D space that fall inside an object~\cite{oechsle2021unisurf}. 
The latter learns a network that predicts the SDF of a 3D point (i.e., its signed distance to the closest surface). 
The surface is extracted from the SDF by finding the zero-level set. 
Our work, inspired by Neuralangelo~\cite{li2023neuralangelo}, builds on a hashgrid-based SDF representation ~\cite{muller2022instant}.
However, unlike in previous work, we use transient measurements as supervision.

Recent works~\cite{yan2023efficient, jungerman2024radiance, qadri2024aoneus, fujimura2023nlos, mirdehghan2024turbosl} have also explored training neural surface representations using data from emerging sensors. 
Concurrent work~\cite{mu2024towards} reconstructed 3D shapes of objects using transient measurements from a low-cost SPAD sensor. 
Their setup differs from ours as they utilize hundreds of training poses, while we investigate 3D surface reconstruction from 2 to 5 viewpoints.
To the best of our knowledge, our work is the first to tackle the problem of few-viewpoint surface reconstruction using transient measurements. 

\paragraph{Few-viewpoint neural reconstruction.}
Prior work has addressed few-viewpoint reconstruction using geometric supervision (e.g., based on depth maps, point clouds, or transients)~\cite{malik2023transient, wang2023sparsenerf, yu2022monosdf, deng2022depth, roessle2022dense}, or by introducing other strategies to regularize reconstruction from multi-view images~\cite{niemeyer2022regnerf,yu2022monosdf, kim2022infonerf, kwak2023geconerf, xu2022sinnerf, yu2021pixelnerf, wu2023s, yang2023freenerf}. Our work is closest to TransientNeRF~\cite{malik2023transient} and RegNeRF~\cite{niemeyer2022regnerf}. 
In contrast to TransientNeRF---which reconstructs a volumetric scene model from few-viewpoint transients---we focus on surface reconstruction using an SDF-based representation and improve robustness to noisy (low photon) measurements. 
Similar to RegNeRF, we regularize views rendered from novel viewpoints; however, we use a different depth-based regularization strategy and exploit transient data to achieve improved performance in the few-view setting.
\section{Method}
\label{sec:method}
 
In this section, we provide a mathematical description of single-photon lidar~\cite{malik2023transient}, which we then relate to the rendering model used to optimize the surface representation. An overview of the method is shown in Figure~\ref{fig:method}.

\subsection{Measurement Model}

Suppose that light from a laser pulse is reflected from a scene and focused onto a sensor at point $\mathbf{p} \in \mathbb{R}^2$. 
Also, assume that laser and sensor are co-axial, such that light propagates to the scene and back along the same path.
Then, we can model the forward path of light along a ray $\mathbf{r}(t) = \mathbf{o}+tc\bm{\omega}(\mathbf{p})$, where $\mathbf{o}\in\mathbb{R}^{3}$ represents the camera origin, $c$ is the speed of light, and $\bm{\omega}$ is a unit vector describing the direction of the ray. 

Light incident along $\mathbf{r}$ is integrated at a sensor pixel $(i, j)$ over time bins $n$, resulting in a transient measurement $\bm{\tilde{\tau}}[i, j, n]$. 
The number of photons collected in each time bin follows a Poisson distribution~\cite{malik2023transient, rapp2017few}:
\begin{align}
    \bm{\tilde{\tau}}[i,j,n] &\sim \text{Poisson}(N\eta \bm{\lambda}[i,j,n] + B), \notag\\ \quad \text{where}\quad B&=N(\eta A[i,j] + D).
\end{align}
Here, $\bm{\lambda}$ is the expected number of incident photons and depends on the reflectivity of the point imaged by pixel $(i, j)$. The value $N$ denotes the number of emitted laser pulses per pixel, $\eta$ is the detection efficiency of the sensor, and $B$ is the number of background detections, which depends on the average ambient photon rate at pixel coordinates $(i,j)$, $A[i,j]$. Finally, the value $D$ is the dark count---i.e., the number of false detections produced by the sensor. 

 For a Poisson distribution, the signal-to-noise ratio (SNR)
 is defined as the ratio between the mean and its standard deviation. Hence, in our case, the SNR of $\tilde{\bm{\tau}}$ is given by:
\begin{align}
      \frac{N\eta \bm{\lambda}[i,j,n] + B}{\sqrt{N\eta \bm{\lambda}[i,j,n] + B}}.
\end{align}
Decreasing the number of laser pulses reduces the acquisition time, the number of collected photons, and the SNR of the measurements.

\subsection{Surface-based Transient Rendering}

We render transients by combining a neural representation that parameterizes the scene appearance and surface geometry with a time-resolved volume rendering equation.

\paragraph{Neural surface representation.}
We use a neural surface representation $\mathcal{F}$ to represent the scene appearance and surface geometry.
The representation consists of a multi-resolution hash-grid, a base multi-layer perceptron (MLP)  $\mathcal{F}_b$,  and a colour MLP  $\mathcal{F}_c$~\cite{muller2022instant,li2023neuralangelo}. 
The MLP $\mathcal{F}_b$ maps a point along a ray $\mathbf{r}(t)\in\mathbb{R}^3$ to a geometric latent feature vector $\bm{l}_{\text{geo}} \in \mathbb{R}^{16}$ and an SDF value $f \in \mathbb{R}$. 
The surface normal is given as $\bm{n} = \frac{\nabla f(\mathbf{r})}{\|\nabla f(\mathbf{r}) \|}$, where the gradient $\nabla f(\mathbf{r})$ is computed via finite differences. 
We then recover the radiance at $\mathbf{r}(t)$ as $\bm{c} = \mathcal{F}_{\bm{c}}(\bm{n}, \gamma(\bm{\omega}),\bm{l}_{\text{geo}})$, where  $\gamma$ is a positional encoding function~\cite{tancik2020fourier}. 

To make the representation compatible with time-resolved volume rendering, we follow NeuS~\cite{wang2021neus} and convert the predicted SDF to a density field $\sigma(\mathbf{r})$ as
\begin{align}
    \sigma(\mathbf{r}) = \max\left(\frac{-\frac{d}{dt}\Phi_s\left(f(\mathbf{r})\right)}{\Phi_s\left(f(\mathbf{r})\right)}, 0\right),
    \label{eq:density}
\end{align}
where $\Phi_s(x)=(1+e^{-sx})^{-1}$ is the sigmoid function, and $s$ is a learnable parameter controlling the support of the density field around the zero level set~\cite{wang2023adaptive}.

\paragraph{Time-resolved volume rendering.} 
Following TransientNeRF~\cite{malik2023transient} we use the predicted density and radiance values from this representation to render transients $\boldsymbol{\tau}$ at each pixel $(i, j)$ and time bin $n$ using the time-resolved volume rendering equation:
\begin{align} \label{eq:transient_rendering}
   \boldsymbol{\tau}[i, j, n] &= \int_{\mathcal{P}_{i, j}} \int_{\mathcal{T}_n} (tc)^{-2}\,T(t)^2\sigma(\mathbf{r})\mathbf{c}(\mathbf{r}, \boldsymbol{\omega}) \,\mathrm{d}t\, \mathrm{d}\mathbf{p}, \\ 
                        \text{where} \,\,\, T(t) &= \exp\left( -\int_{t_0}^{t}\sigma(\mathbf{r})\,\mathrm{d}s\right)\nonumber.
\end{align}
The value $T(t)$ is the transmittance from a distance $t_0$ along the ray to $t$, and this term is squared to reflect the two-way propagation of light~\cite{attal2021torf}.
The value $(tc)^{-2}$ accounts for the inverse-square falloff of intensity, $\mathcal{T}_n$ is the transient bin width, and $\mathcal{P}_{i, j}$ models the pixel footprint. Note that in practice, we approximate the integral using quadrature by sampling at locations $t_i$ along the ray~\cite{max1995optical}. We approximate the integral over $\mathcal{P}_{i, j}$ by explicitly sampling multiple rays within the area of each pixel. Finally, we convolve $\boldsymbol{\tau}$ with the calibrated temporal impulse response of the lidar system to yield the final rendered transient $\boldsymbol{\tau}_f$~\cite{malik2023transient}.

\subsection{Optimization}

\paragraph{Data term.}
We supervise the rendered transients to be consistent with the measurements $\mathbf{\tilde{\bm{\tau}}}$ by minimizing the following loss function.
\begin{align}\label{eq:transient_loss}
  \mathcal{L}_{\bm{\tau}} = \frac{1}{|\mathcal{R}_\text{train}|}\sum_{\mathcal{R_\text{train}}}\|\tilde{\bm{\tau}} - {\bm{\tau}_f}\|_{1},
\end{align}
where $\mathcal{R}_\text{train}$ is the set of rays corresponding to the captured transient measurements used for training.

\paragraph{Weight variance regularization.} 
To recover thinner and smoother surfaces and aid the model in generalizing to viewpoints far from the training distribution, 
we regularize the scene density from sampled unseen viewpoints by penalizing the variance of the rendering weights along each ray.

To sample unseen viewpoints, we follow RegNeRF~\cite{niemeyer2022regnerf} and sample camera origins on a sphere around the scene estimated from the training views (see supplement for details). 
We then regularize the density from these unseen views using RawNeRF's weight-variance regularizer~\cite{mildenhall2022raw}. 
We find that replacing their rendered depth estimate with the argmax depth~\cite{malik2023transient} recovers more accurate surfaces, as shown in the supplement.

More specifically, let $t_i$ be the samples along the ray used in the quadrature approximation of Eq.~\ref{eq:transient_rendering}. Then, the values $w_i = \sigma(t_i)T(t_i)$ represent alpha compositing weights applied to the radiance values along a ray. 
We use $w_i$ to compute the weight-variance regularizer, which we apply to rays $\mathcal{R}_\text{train}^{\prime}$ sampled from the set of unseen views:

\begin{align}
\mathcal{L}_{\text{weight\_var}} = \frac{1}{|\mathcal{R}_\text{train}^\prime|} \sum_{\mathcal{R}_\text{train}^{\prime}}
\sum_{i} w_i \frac{(t_i - d)^3 + (t_{i-1} - d)^3}{3 (t_i - t_{i-1})}. 
\label{eq:weight_var}
\end{align}
The value $d = \argmax_{t^*}\; T(t^*)\sigma(t^*)$ represents the depth, or the distance along the ray with the maximum probability of ray termination~\cite{malik2023transient}.

\paragraph{Reflectivity loss.}

To improve performance in the low photon count regime, we introduce a reflectivity loss, which provides additional supervision using the integrated transients. 
Since the integrated transients have a higher signal-to-noise ratio than the individual photon count histogram bins, this provides a complementary source of supervision.
We implement this loss as
\begin{align}\label{eq:Reflectivity loss}
\mathcal{L}_{\text{ref}} = \frac{1}{|\mathcal{R}_\text{train}|}\sum_{\mathcal{R_\text{train}}}\|\sum_{n}\bm{\tau}[i,j,n]-\sum_{n}\tilde{\bm{\tau}}[i,j,n]\|_1,
\end{align}
where the integrated transients are computed by summing over $n$. 

In summary, our training loss can be expressed as
\begin{align}\label{eq:training_loss}
  \mathcal{L} &= \mathcal{L}_{\bm{\tau}} + \lambda_{\text{ref}} \mathcal{L}_{\text{ref}} + \lambda_{\text{sc}} \mathcal{L}_{\text{sc}} + \lambda_{\text{eik}} \mathcal{L}_{\text{eik}} \nonumber \\
  &\quad + \lambda_{\text{weight\_var}} \mathcal{L}_{\text{weight\_var}} + \lambda_{\text{sparse}} \mathcal{L}_{\text{sparse}}.
\end{align}
Here, the space carving loss, $\mathcal{L}_\text{sc}$, penalizes densities at locations where the transient is below a background level $B$: 
\begin{align}\label{eq:space carving loss}
 \mathcal{L}_{\text{sc}} = \sum\limits_{\substack{i, j, n \\ \boldsymbol{\widetilde{\tau}}[i, j, n] < B}} \int_{\mathcal{P}_{i, j}} \int_{\mathcal{T}_n} T(t)\sigma(\mathbf{r}) \,\mathrm{d}t\, \mathrm{d}\mathbf{p}.
\end{align}
The Eikonal regularization~\cite{gropp2020implicit} is given as 
\begin{align}\label{eq:eikonal loss}
 \mathcal{L}_{eik} = \frac{1}{|\mathbf{r} \in \mathcal{R}_\text{train}|} \sum_{ \mathbf{r} \in \mathcal{R}_\text{train}}(\|\nabla f(\mathbf{r})\| - 1)^2,
\end{align}
and encourages the SDF to have a gradient magnitude equal to one.
Finally, the sparsity loss is
\begin{align}\label{eq:sparsity loss}
 \mathcal{L}_{\text{sparse}} = \frac{1}{|\mathbf{r} \in \mathcal{R}_\text{train}|} \sum_{ \mathbf{r} \in \mathcal{R}_\text{train}} \exp(-\alpha |f(\mathbf{r})|),
\end{align}
where $\alpha$ is a scalar hyperparameter. This loss helps to prevent floating artifacts in the reconstruction.

\subsection{Implementation Details}
\label{sec:imp_detials}
We implement our method in Pytorch-Lightning~\cite{falcon2019pytorch, instant-nsr-pl}. 
We use the NerfAcc~\cite{li2022nerfacc} version of InstantNGP \cite{muller2022instant} for efficient ray marching, and we build on top of TransientNeRF's~\cite{malik2023transient} time-resolved rendering.

\paragraph{Training and optimization.}
We initialize the SDF to approximate a sphere \cite{atzmon2020sal}. We follow Neuralangelo's coarse-to-fine optimization strategy by initializing the step-size in the gradient calculation $\epsilon$ as the coarsest hash-grid voxel size and exponentially decreasing it during training. For the hash-grid, we start at four hash resolutions and unmask two new hash resolutions every 5000 steps. We optimize the network using the AdamW~\cite{loshchilov2017decoupled} optimizer, with $\beta_{1} = 0.9$ and $\beta_{2} = 0.99$ and employ a two-step learning rate scheduler by initializing the learning rate to $1\times10^{-5}$ and linearly increasing it until it reaches $1\times10^{-3}$ by the end of the 5000 steps. Then, the learning rate is exponentially decreased over the remaining steps, with a total of 250K steps. All experiments were run on a single Nvidia RTX A6000 GPU with 48GB of memory. We provide additional implementation details, including hyperparameter settings, in the supplement.

\section{Multiview Transient Dataset}

To test the robustness of different methods on low-SNR transient data, we adapt the simulated and captured datasets introduced by TransientNeRF~\cite{malik2023transient}. For each captured and simulated scene, we synthesize transients with an average of 300, 150, 50, and 10 photons per occupied (non-background) pixel. The paragraphs below outline how these datasets are created. 
\vspace{-1em}
\paragraph{Simulated dataset.} To synthesize transients at different photon counts in simulation, we match the flux from the time-resolved Mitsuba renderer~\cite{malik2023transient} to the target number of photons per pixel. We then follow Malik et al.~\cite{malik2023transient} and set the background flux to $0.001$ photons per $2850$ captured scene photons, which is set to match the statistics of the captured dataset. The flux is then Poisson sampled to simulate photon counts.
\vspace{-1em}
\paragraph{Captured dataset.} 
We create low-flux captured datasets using the raw photon arrival timestamps from TransientNeRF. Specifically, we subsample the single photon arrivals using thinning \cite{lewis1979simulation} to achieve a set number of detected photons. We then bin the photon arrivals into a transient histogram.
Additional dataset details can be found in the supplement. 
Both the simulated and captured variable photon count datasets will be released upon publication.

\section{Results}
\label{sec:results}

\begin{figure*}[t]
    \centering
    \includegraphics[width=\textwidth]{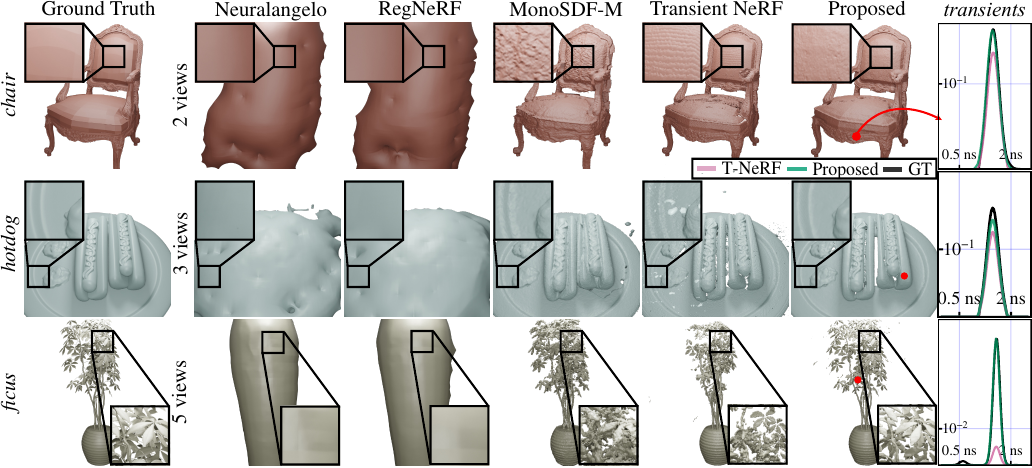}
    \caption{Results on the simulated dataset using an average of 6000 photons per occupied (non-background) pixel. We show the recovered meshes from the baselines and the proposed method. For the transient-based methods, we also show rendered transients for the indicated pixels. Our method recovers smoother meshes with fewer missing parts. We also recover transients that better match the ground truth.}
    \label{fig:res_sim}
\end{figure*}

\begin{table*}
    \caption{Simulated results assessing image quality, depth accuracy, and mesh quality. The left column indicates average photons per pixel. }
    \label{tab:simulated}
    \centering
    \resizebox{\textwidth}{!}{%
    \begin{tabular}{ll|ccc|ccc|ccc|ccc|ccc}
        \toprule  
        & & \multicolumn{3}{c|}{Chamfer Distance $\,\downarrow$} & \multicolumn{3}{c|}{PSNR (dB)$\,\uparrow$} & \multicolumn{3}{c|}{LPIPS$\,\downarrow$} & \multicolumn{3}{c|}{L1 (depth) $\,\downarrow$} & \multicolumn{3}{c}{Transient IOU$\,\uparrow$}\\ 
        & \textbf{Method}                             & 2 views & 3 views & 5 views & 2 views & 3 views & 5 views & 2 views & 3 views & 5 views  & 2 views & 3 views & 5 views & 2 views & 3 views & 5 views\\\midrule
              \parbox[t]{6mm}{\multirow{5}{*}{\rotatebox[origin=c]{90}{\textit{6000}}}} 

        & Neuralangelo~\cite{li2023neuralangelo}        & 4.56 & 7.99 & 7.17 & 18.57 & 19.44 & 22.03 & 0.388 & 0.442 & 0.434 & 0.162 & 0.122 & 0.090 & - & - & -\\

        & RegNeRF ~\cite{niemeyer2022regnerf}          & 4.94 & 3.48 & 4.38 & 18.32 & 19.89 & 21.67 & 0.376 &  0.412 &  0.418 & 0.264 & 0.142 & 0.100 & - & - & - \\
        & MonoSDF w/ mask ~\cite{yu2022monosdf}       & \cellcolor{yellow}3.78 & \cellcolor{yellow}1.67 & \cellcolor{orange}0.18 & \cellcolor{yellow}19.76 & \cellcolor{yellow}20.74 & \cellcolor{yellow}24.65 & \cellcolor{yellow}0.240 & \cellcolor{yellow}0.244 & \cellcolor{orange}0.132 & \cellcolor{yellow}0.045 &\cellcolor{yellow} 0.026 & \cellcolor{yellow}0.015 & - & - & - \\
        & TransientNeRF \cite{malik2023transient}   & \cellcolor{orange}0.27 & \cellcolor{orange}0.44 & \cellcolor{yellow}0.31 & \cellcolor{orange}21.38 & \cellcolor{orange}23.48 & \cellcolor{orange}28.39 & \cellcolor{red}0.172 & \cellcolor{red}0.151 & \cellcolor{red}0.115 & \cellcolor{orange}0.015 & \cellcolor{orange}0.011 & \cellcolor{orange}0.013 & \cellcolor{orange}0.31 & \cellcolor{orange}0.40 & \cellcolor{orange}0.55\\
        & Proposed                                    & \cellcolor{red} 0.07 & \cellcolor{red}0.06 & \cellcolor{red} 0.05 & \cellcolor{red} 25.28 & \cellcolor{red} 26.69 & \cellcolor{red}29.49 & \cellcolor{orange}0.176 & \cellcolor{orange}0.155 & \cellcolor{yellow}0.150 & \cellcolor{red}0.006 & \cellcolor{red}0.006 & \cellcolor{red}0.005 & \cellcolor{red}0.54 & \cellcolor{red}0.58 & \cellcolor{red}0.68\\

        \midrule
        
              \parbox[t]{6mm}{\multirow{2}{*}{\rotatebox[origin=c]{90}{\textit{300}}}} 
        & TransientNeRF \hspace{1em}     & 0.26 & 0.33 & 0.30 & 21.34 & 22.79 & 25.88 & 0.201 & \cellcolor{red}0.152 & \cellcolor{red}0.121 & 0.012 & 0.012 & 0.015 & 0.32 & 0.41 & 0.58\\

       & Proposed \hspace{1em}          & \cellcolor{red}0.07 & \cellcolor{red}0.06 & \cellcolor{red}0.06 & \cellcolor{red}25.70 & \cellcolor{red}26.65 & \cellcolor{red}28.50 & \cellcolor{red}0.196 & 0.162 &         0.149 & \cellcolor{red}0.007 &\cellcolor{red} 0.007 &\cellcolor{red} 0.005 &\cellcolor{red} 0.56 &\cellcolor{red} 0.60 &\cellcolor{red} 0.70 \\
        \midrule
        \parbox[t]{6mm}{\multirow{2}{*}{\rotatebox[origin=c]{90}{\textit{150}}}} 
        
        & TransientNeRF  & 0.25 & 0.65 & \cellcolor{red}0.27 & 21.04 & 22.18 & 25.35 & 0.203 & \cellcolor{red}0.161 & \cellcolor{red}0.117 & 0.015 & 0.013 & 0.010 & 0.31 & 0.39 & 0.56\\
        & Proposed  & \cellcolor{red}0.09 & \cellcolor{red}0.07 & 0.29 & \cellcolor{red}25.56 & \cellcolor{red}26.52 &\cellcolor{red} 28.19 &\cellcolor{red} 0.193 & 0.165 & 0.151 &\cellcolor{red} 0.007&\cellcolor{red} 0.006& \cellcolor{red}0.005& \cellcolor{red}0.56 & \cellcolor{red}0.60 & \cellcolor{red}0.70 \\
        \midrule
        \parbox[t]{6mm}{\multirow{2}{*}{\rotatebox[origin=c]{90}{\textit{50}}}} 
        &TransientNeRF & 0.53 & 0.66 & 0.50 & 20.16 & 21.57 & 24.01 & 0.232 & 0.184 & \cellcolor{red}0.133 & 0.021 & 0.017 & 0.014 & 0.29 & 0.40 & 0.55\\
        & Proposed & \cellcolor{red}0.15 & \cellcolor{red}0.08 & \cellcolor{red}0.06 & 2\cellcolor{red}4.73 &\cellcolor{red} 25.76 &\cellcolor{red} 26.55 & \cellcolor{red}0.218 &\cellcolor{red} 0.163 & 0.167 & \cellcolor{red}0.012 &\cellcolor{red} 0.005 &\cellcolor{red} 0.007 &\cellcolor{red}0.53 &\cellcolor{red}0.59 &\cellcolor{red}0.63 \\
        \midrule
              \parbox[t]{6mm}{\multirow{2}{*}{\rotatebox[origin=c]{90}{\textit{10}}}} 
        & TransientNeRF \hspace{1em}     & 2.86 & 0.31 & 5.50 & 18.94 & 22.85 & 16.73 & 0.265 & 0.218 & 0.209 & 0.052 & 0.018 & 0.092 & 0.27 & 0.43 & 0.19 \\

       & Proposed \hspace{1em}          & \cellcolor{red}0.22 & \cellcolor{red}0.28 & \cellcolor{red}0.18 & \cellcolor{red}22.76 & \cellcolor{red}23.98 & \cellcolor{red}25.51 & \cellcolor{red}0.254 & \cellcolor{red}0.216 & \cellcolor{red}0.208 & \cellcolor{red}0.016 &\cellcolor{red} 0.015 &\cellcolor{red} 0.014 &\cellcolor{red} 0.43 &\cellcolor{red} 0.47 &\cellcolor{red} 0.54 \\
        \bottomrule
\end{tabular}}
\end{table*}

\begin{figure*}[t]
    \centering
    \includegraphics[width=\textwidth]{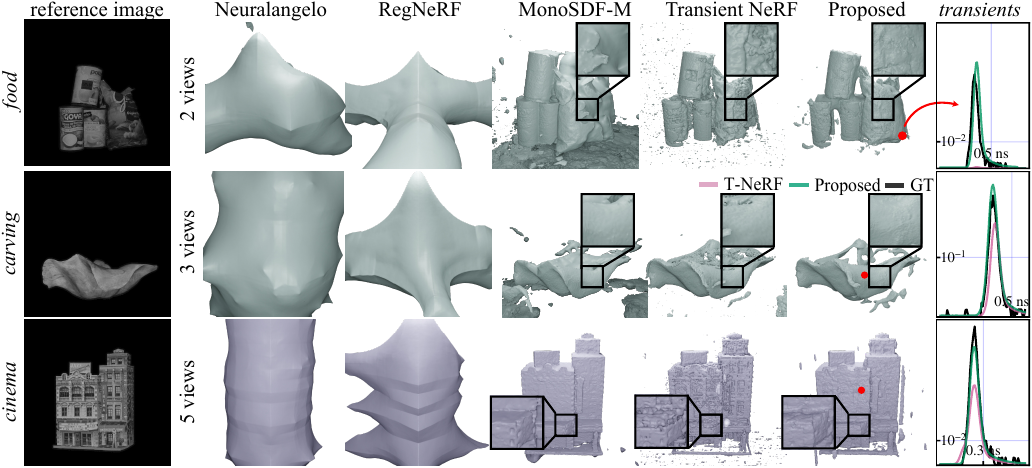}
    \caption{Results on the captured dataset. We show the recovered meshes from the baselines and the proposed method. Due to the lack of ground-truth, we include the closest captured image for reference of the scene. As can be seen, Neuralangelo recovers smoother meshes with fewer missing parts. We also recover transients that better match the ground truth.}
    \label{fig:res_cap}
\end{figure*}

\begin{table*}
    \caption{Captured results assessing image quality, depth accuracy, and mesh quality. The left column indicates average photons per pixel.}
    \label{tab:captured}
    \centering
    \resizebox{\textwidth}{!}{%
    \begin{tabular}{ll|ccc|ccc|ccc|ccc}
        \toprule  
         & & \multicolumn{3}{c|}{PSNR (dB)$\,\uparrow$} & \multicolumn{3}{c|}{LPIPS$\,\downarrow$} & \multicolumn{3}{c|}{L1 (depth) $\,\downarrow$} & \multicolumn{3}{c}{Transient IOU$\,\uparrow$} \\ 
        & \textbf{Method}                            &  2 views & 3 views & 5 views & 2 views & 3 views & 5 views  & 2 views & 3 views & 5 views & 2 views & 3 views & 5 views\\\midrule
        \parbox[t]{6mm}{\multirow{5}{*}{\rotatebox[origin=c]{90}{\textit{1500}}}} 
        &Neuralangelo~\cite{li2023neuralangelo}        &17.79 & 19.28 & 21.07 & 0.380 & 0.355 & 0.300 & 0.076 & 0.071 & 0.061 & - & - & - \\

        &RegNeRF ~\cite{niemeyer2022regnerf}           &\cellcolor{yellow}18.50 & 19.59 & 20.29 & 0.375 & 0.320 & 0.330 & 0.085 & 0.073 & 0.076 & - & - & - \\
    
        &MonoSDF w/ mask~\cite{rematas2022urban}   &17.63 & \cellcolor{yellow}21.16 & \cellcolor{red}27.25 & \cellcolor{yellow}0.353 & \cellcolor{yellow}0.248 & \cellcolor{orange}0.157 & \cellcolor{yellow}0.033 & \cellcolor{yellow}0.026 & \cellcolor{yellow}0.021 & - & - & -\\
        &TransientNeRF \cite{malik2023transient}   & \cellcolor{red}22.11 & \cellcolor{orange}21.83 & \cellcolor{yellow}22.72 &\cellcolor{orange} 0.271 & \cellcolor{orange}0.212 &\cellcolor{yellow} 0.172 & \cellcolor{red}0.005 & \cellcolor{red}0.006 & \cellcolor{orange}0.010 & \cellcolor{orange}0.34 & \cellcolor{orange}0.41 & \cellcolor{orange}0.50\\
        &Proposed                                    & \cellcolor{orange}21.31 & \cellcolor{red}23.95 & \cellcolor{orange}25.22 & \cellcolor{red}0.241 & \cellcolor{red}0.170 & \cellcolor{red}0.145 & \cellcolor{orange}0.007 & \cellcolor{red}0.006 & \cellcolor{red}0.006 & \cellcolor{red}0.40 & \cellcolor{red}0.50 & \cellcolor{red}0.58 \\
        \midrule
        
        \parbox[t]{6mm}{\multirow{2}{*}{\rotatebox[origin=c]{90}{\textit{300}}}} 
        &TransientNeRF      & 20.14 & 21.99 & 20.53 & 0.270 & 0.210 & 0.177 & 0.012 & 0.007 & 0.014 &0.26 &0.39 & 0.36 \\
        &Proposed      & \cellcolor{red}21.60 & \cellcolor{red}23.53 & \cellcolor{red}23.79 & \cellcolor{red}0.223 & \cellcolor{red}0.170 & \cellcolor{red}0.151 & \cellcolor{red}0.005 & \cellcolor{red}0.005 & \cellcolor{red}0.012 & \cellcolor{red}0.42 & \cellcolor{red}0.49 & \cellcolor{red}0.55 \\
        \midrule
        \parbox[t]{6mm}{\multirow{2}{*}{\rotatebox[origin=c]{90}{\textit{150}}}} 
        &TransientNeRF \hspace{1em}    & 19.13 & 21.06 & 19.95 & 0.277 & 0.206 & 0.196 &0.016 & 0.015 & 0.018 & 0.21 & 0.33 & 0.28 \\
        &Proposed \hspace{1em}         & \cellcolor{red}21.23 &\cellcolor{red} 23.56 & \cellcolor{red}25.33 & \cellcolor{red}0.222 & \cellcolor{red}0.171 & \cellcolor{red}0.146& \cellcolor{red}0.007 & \cellcolor{red}0.005 &\cellcolor{red}0.004 & \cellcolor{red}0.41 & \cellcolor{red}0.50 & \cellcolor{red}0.60\\                
        \midrule
        \parbox[t]{6mm}{\multirow{2}{*}{\rotatebox[origin=c]{90}{\textit{50}}}} 
        
        &TransientNeRF \hspace{1em}      & 18.80 & 19.63 & 19.31 & \cellcolor{red}0.274 & 0.194 & 0.190 & 0.020 & 0.020 & 0.018 & 0.19 & 0.26 & 0.24 \\
        &Proposed \hspace{1em}           & \cellcolor{red}19.14 & \cellcolor{red}22.33 & \cellcolor{red}25.34 & 0.280 & \cellcolor{red}0.180 & \cellcolor{red}0.148 & \cellcolor{red}0.017 & \cellcolor{red}0.011 &\cellcolor{red}0.007 & \cellcolor{red}0.34 & \cellcolor{red}0.44 & \cellcolor{red}0.59 \\

        \midrule
        \parbox[t]{6mm}{\multirow{2}{*}{\rotatebox[origin=c]{90}{\textit{10}}}} 
        
        &TransientNeRF \hspace{1em}      & 16.40 & 16.36 & 15.40 & 0.246 & 0.259 & 0.272 & 0.037 & 0.041 & 0.090 & 0.08 & 0.08 & 0.02 \\
        &Proposed \hspace{1em}           & \cellcolor{red}22.40 & \cellcolor{red}23.05 & \cellcolor{red}25.47 & \cellcolor{red}0.242 & \cellcolor{red}0.208 & \cellcolor{red}0.173 & \cellcolor{red}0.008 & \cellcolor{red}0.008 &\cellcolor{red}0.007 & \cellcolor{red} 0.35 & \cellcolor{red}0.36 & \cellcolor{red}0.45 \\
        \bottomrule
\end{tabular}}
\end{table*}

We evaluate our method for transient and intensity image synthesis from novel viewpoints, depth rendering, and mesh recovery. 
We also benchmark the performance of our method's robustness to low-SNR data. Our supplementary contains a more detailed set of results and ablation studies.

\paragraph{Baselines.} 
We compare the rendered intensity, depth maps, and meshes against state-of-the-art methods in surface reconstruction and few-view surface reconstruction: Neuralangelo~\cite{li2023neuralangelo}, RegNeRF~\cite{niemeyer2022regnerf}, and MonoSDF~\cite{yu2022monosdf}. 
To ensure a fair comparison, we implement these baselines using the same Neuralangelo~\cite{li2023neuralangelo} backbone architecture, and we integrate their proposed loss functions. 
All three methods are supervised using integrated transient measurements. 
MonoSDF is also supervised using depth maps obtained by applying a log-matched filter~\cite{kirmani2014first} to the transient measurements, and the input normal maps are estimated from the depth. 
We also compare our method to TransientNeRF~\cite{malik2023transient}. 

\paragraph{Evaluation criteria.} 
To measure the mesh reconstruction quality, we use Chamfer Distance. For image reconstruction evaluation, we use PSNR and LPIPS \cite{zhang2018unreasonable}. We also evaluate the quality of the rendered transients using Transient IOU~\cite{malik2023transient}, defined as the intersection over the union of the rendered transient and the ground truth transient (see supplement for details).

\subsection{Simulated Results}
Figure \ref{fig:res_sim} shows the recovered meshes and rendered transients for specific pixels from the proposed method compared to baselines. 
We show results for measurements with an average of 6000 photons per occupied pixel (results for other photon levels are included in the supplement). 
Due to a lack of any depth-based priors, Neuralangelo~\cite{li2023neuralangelo} and RegNeRF~\cite{niemeyer2022regnerf} both fail to recover accurate scene geometry. MonoSDF~\cite{yu2022monosdf} recovers noisier surfaces than the proposed approach, especially when trained with fewer input viewpoints. Compared to TransientNeRF~\cite{malik2023transient}, the proposed method recovers smoother surfaces with fewer missing parts and fewer floating artifacts, especially in the case of training with two or three input viewpoints. 
The results highlight the benefit of training a neural surface representation with transient supervision.

\begin{figure*}[th!]
    \centering
    \includegraphics[width=\textwidth]{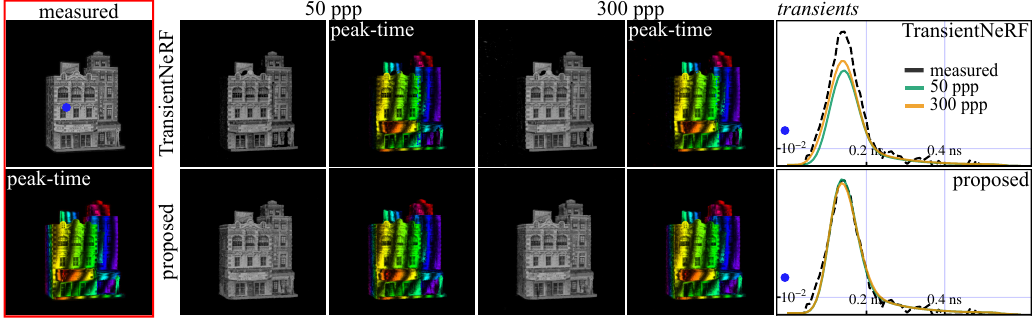}
    \caption{Novel view synthesis for varying photon levels. We show the rendered novel view on the \textit{cinema} scene, trained on five viewpoints with an average of $50, 300$ photons.  We also show peak-time visualizations~\cite{malik2024flying}, which show the full transient in a single visualization. Hue encodes the
time of peak intensity, brightness is modulated by the maximum intensity, and each
band corresponds to an isochrone, or wavefront of equal path length. We show transient plots (right) for the pixel indicated on the ground-truth image (blue dot).}
    \label{fig:res_lp}
\end{figure*}

Table~\ref{tab:simulated} shows the quantitative performance of the proposed method against the baselines averaged across all scenes in the two, three, and five training viewpoint settings. 
We benchmark performance for varying measurement SNRs by training using transients with lower photon counts. 
To compute the metrics, we first scale the rendered transients so that the average photon count per occupied pixel (across all viewpoints) is the same across all photon levels.
The recovered meshes from the proposed method achieve significantly lower Chamfer Distance compared to the other methods and roughly $5 \times$ lower than the second-best method, TransientNeRF. 
The recovered depth maps are also more accurate than the baseline methods, and the image metrics are comparable. 
Finally, the transients rendered from our method are more similar to the ground truth based on the Transient IOU metric.

\renewcommand{\thefootnote}{\fnsymbol{footnote}}
\begin{table}[h!]
    \caption{Ablation study on the \textit{Lego} scene with 6000ppp. We present results without the sparsity loss (w/o $\mathcal{L}_{\text{sparse}}$), without the space carving loss (w/o $\mathcal{L}_{\text{sc}}$), without the weight variance regularizer (w/o $\mathcal{L}_{\text{weight\_var}}$) and without the reflectivity loss (w/o $\mathcal{L}_{\text{ref}}$).} 
    \label{tab:ablation_loss}
    \vspace{-1em}
    \begin{center}
    \resizebox{0.47\textwidth}{!}{%
    \addtolength{\tabcolsep}{-0.43em}
    \begin{tabular}{l|ccc|ccc|ccc}
        \toprule  
         &\multicolumn{3}{c|}{Chamfer Distance $\,\downarrow$} & \multicolumn{3}{c|}{PSNR (dB)$\,\uparrow$} & \multicolumn{3}{c}{Transient IOU$\,\uparrow$}\\ 
         \textbf{Method}                             & 2 views & 3 views & 5 views & 2 views & 3 views & 5 views  & 2 views & 3 views & 5 views \\\midrule

        Ours w/o $\mathcal{L}_{\text{sc}}$          & \cellcolor{tabsecond}0.07 & \cellcolor{yellow}0.04 & \cellcolor{tabthird}0.11 & \cellcolor{tabsecond}25.63 & \cellcolor{tabthird}26.32 & \cellcolor{tabthird}26.22 & \cellcolor{tabfirst}0.53 & \cellcolor{tabsecond}0.57 & 0.53 \\
        Ours w/o $\mathcal{L}_{\text{weight\_var}}$ & 0.28 & 0.37 & 1.04 & 21.58 & 22.66 & 23.81 & \cellcolor{tabsecond}0.48 & 0.53 & \cellcolor{tabthird}0.60 \\
        Ours w/o $\mathcal{L}_{\text{sparse}}$      & \cellcolor{tabfirst}0.02 & \cellcolor{tabsecond}0.03 & \cellcolor{tabsecond}0.08 & \cellcolor{tabthird}25.43 & \cellcolor{tabfirst}26.87 & \cellcolor{tabsecond}26.58  & \cellcolor{tabfirst}0.53 & \cellcolor{tabfirst}0.58 & \cellcolor{tabsecond}0.63\\
        Ours w/o $\mathcal{L}_{\text{ref}}* $  &  5.50& 2.91 & 0.53 & 22.38 & 24.92  & \cellcolor{tabthird}26.22 & \cellcolor{tabthird}0.49 & \cellcolor{tabthird}0.55 & \cellcolor{tabfirst}0.65\\
                Ours                             & \cellcolor{tabfirst}0.02 & \cellcolor{tabfirst}0.02 & \cellcolor{tabfirst}0.01 & \cellcolor{tabfirst}25.72 & \cellcolor{tabsecond}26.66 & \cellcolor{tabfirst}27.47  & \cellcolor{tabfirst}0.53 & \cellcolor{tabfirst}0.58 & \cellcolor{tabsecond}0.63 \\
        \bottomrule
    \end{tabular}}
    \end{center}
    \vspace{-1em}
    {\scriptsize *For this ablation we set the Eikonal loss to $0$ and the weight variance loss to $10^{-5}$.}
\end{table}


\subsection{Captured Results}
Figure \ref{fig:res_cap} shows the recovered meshes for our method and the baselines on the \textit{food, carving, cinema} scenes for two, three, and five training viewpoints respectively (see the supplement for qualitative results for all scenes). As in TransientNeRF~\cite{malik2023transient}, we mask out the background of the transients for the captured dataset. Similarly to the simulated dataset, the meshes recovered from the proposed method are more accurate than the baseline methods.

We report metrics averaged across six captured scenes in Table~\ref{tab:captured}. Due to a lack of ground-truth meshes, we are not able to report Chamfer distances, and the depth L1 metric is calculated by comparing depth maps estimated from log-matched filtering of transients from held-out views. Our quantitative results show an improvement over TransientNeRF across all image and depth metrics. Overall, our method remains robust across various photon levels, whereas TransientNeRF's image rendering quality deteriorates rapidly as the mean photon count decreases.

Figure~\ref{fig:res_lp} shows the results of training our method and TransientNeRF~\cite{malik2023transient} on transients with an average of $50$ and $300$ photons per occupied pixel.
At these photon levels, using TransientNeRF results in rendered transients that underestimate the measured intensities compared to our method.

\subsection{Ablation Study}
Table~\ref{tab:ablation_loss} evaluates the effect of ablating the regularization terms on the simulated \textit{Lego} scene. The results indicate that all the added regularizers are crucial for performance; however, $\mathcal{L}_{\text{ref}}$ and $\mathcal{L}_{\text{weight\_var}}$ are the most essential. 
When ablating the reflectivity loss (row 4), we deviated from the default settings of $\mathcal{L}_{\text{eik}}$ and $\mathcal{L}_{\text{weight\_var}}$, as indicated in the table, to improve convergence  (see details in the supplement).

\section{Discussion}
We show that using a neural surface representation with transient supervision and additional regularization recovers high-fidelity surfaces in the few viewpoint setting with as few as 10 measured photons per pixel. 
Our work improves the practicality of transient-based reconstruction by reducing the required light levels, which could be especially useful for 3D imaging of photo-sensitive materials or long-range (and low-SNR) remote sensing.

In future work, we hope to address certain limitations of our method. For example, computational efficiency could potentially be improved by developing new time-resolved rendering frameworks based on Gaussian splatting~\cite{kerbl3Dgaussians}.
Additionally, while the main focus of our work is scene reconstruction using the direct component of light, transients carry much richer scene information, including higher order bounces of light, future work could exploit indirect light transport to infer scene properties such as geometry, reflectance, and material characteristics. 

\section{Acknowledgments}
DBL acknowledges support of the Canada Foundation for Innovation, the Ontario Research Fund, and NSERC under the RGPIN, RTI, and Alliance programs.

{\small
\bibliographystyle{plain}
\bibliography{egbib}
}
\clearpage

\section{Implementation Details}
\paragraph{Network implementation.} In this section, we provide a detailed description of the network architecture. We implement our implicit surface representation network using the open-sourced instant-nsr-pl~\cite{instant-nsr-pl}, which is built on Pytorch Lightning~\cite{falcon2019pytorch}. We combine this with TransientNeRF's~\cite{malik2023transient} neural transient renderer, which extends NerfAcc~\cite{li2022nerfacc} and Instant-NGP~\cite{muller2022instant}. 

All network hyperparameters are shared across our methods and the baseline methods unless stated otherwise. Our network architecture follows closely that of NeuS~\cite{wang2021neus} and Neuralangelo~\cite{li2023neuralangelo}. In particular, the number of hash feature grids is set to 16 with a feature size of $2$, and hash-map size $2^{19}$. 
The base MLP consists of a single hidden layer with $64$ neurons with ReLU activation, which predicts the latent vector and SDF. The color MLP consists of two hidden layers with $64$ neurons each, mapping the normal vector, latent vector, and positionally encoded~\cite{mildenhall2021nerf} viewing direction to color. We further employ the occupancy grid from NerfAcc with a resolution of $128^3$, where the grid is binarized using an occupancy value of $10^{-3}$. We set the bounding box radius to $1.5$ for the simulated dataset and $0.4$ for the captured dataset for all methods.

\subsection{Baseline Implementation Details}
We compare our method against four baseline methods: Neuralangelo~\cite{li2023neuralangelo}, RegNeRF~\cite{niemeyer2022regnerf}, MonoSDF-M~\cite{yu2022monosdf}, and TransientNeRF~\cite{malik2023transient}. To ensure a fair comparison, we implement those methods using NerfAcc and InstantNGP. As the first three methods are trained on images, we create intensity images by integrating the transients over the time dimension, and normalizing them between $[0,1]$ using a scale factor, which is set per scene. Finally, the scaled images are gamma corrected with $\gamma = 2.2$. All image-based baseline methods (Neuralangelo, RegNeRF, MonoSDF-M) are trained using the conventional photometric loss \cite{mildenhall2021nerf}.
\begin{align}\label{eq:Photometric loss}\tag{S1}
\mathcal{L}_{\text{photo}} = \frac{1}{|\mathcal{R}_\text{train}|}\sum_{\mathcal{R}_\text{train}}||\tilde{\bm{C}}(\bm{r}) - \bm{C}(\bm{r}) ||^2_{2}
\end{align}
where $\tilde{\bm{C}}$ and $\bm{C}$ denote the predicted and measured color, respectively. We compute this loss over the set of \emph{active rays}, i.e. the set of rays with non-zero opacity values. \\ 

\noindent On the other hand, the transient-based method TransientNeRF is trained using the HDR-informed loss function:
\begin{align}\label{eq:Transient loss}\tag{S2}
\mathcal{L}_{\bm{\tau}} = \frac{1}{|\mathcal{R}_\text{train}|}\sum_{\mathcal{R_\text{train}}}\|\ln(\tilde{\bm{\tau}}+1) - \ln({\bm{\tau}_f} + 1)\|_{1},
\end{align}
where $\tilde{\bm{\tau}}$ is the measured transient and $\bm{\tau}_f$ is predicted transient.

\paragraph{Neuralangelo.} We train Neuralangelo using only the Eikonal loss~\cite{gropp2020implicit}, as we did not notice any significant improvement by adding the curvature loss. The total objective is $\mathcal{L} = \mathcal{L}_{\text{photo}} + \lambda_{\text{eik}}\mathcal{L}_{\text{eik}}$. For both the simulated and captured datasets, we use $\lambda_{\text{eik}} = 0.1$

\paragraph{RegNeRF.} For RegNeRF we extend Neuralangelo with an additional total variation regularizer on the depths rendered from the unseen cameras. We omit the color regularization as we do not have access to the weights of RealNVP~\cite{dinh2016density}. We also implement Sample Space Annealing over the first 256 iterations. To apply their patch-based regularization, given an existing set of poses $\left\{ \bm{P} \right\}_i = \left\{\bm{R}|\bm{t}\right\}_i$, we calculate the mean focus point by finding the point with shortest distances (in the least squares sense) to the optical axes of those cameras. Then, we calculate the bounding sphere of plausible cameras to sample from by computing the average of the distances of the focus point to the cameras. At each optimization step, we sample a new camera origin from that bounding sphere and align its rotation matrix such that it looks at the mean focus point. We then render an $8\times8$ patch over one of the training images and render the depth $d$ of that patch following NeRF~\cite{mildenhall2021nerf}. The final loss function extends Neuralangelo with the total variation regularizer on the rendered depth:
\begin{align}\label{eq:Total variation loss}\tag{S3}
\mathcal{L}_{\text{ds}} = \sum_{\mathcal{R_\text{sampled}}}\sum_{i,j=1}^{8}({d}(\bm{r}_{ij}) - {d}(\bm{r}_{i+1j}))^2 + {d}(\bm{r}_{ij}) - {d}(\bm{r}_{ij+1}))^2,
\end{align}
where $\mathcal{R_\text{sampled}}$ are the rays from the sampled cameras. Therefore, the total loss is $\mathcal{L} = \mathcal{L}_{\text{photo}} + \lambda_{\text{eik}}\mathcal{L}_{\text{eik}} + \lambda_{\text{ds}}\mathcal{L}_{\text{ds}}$. For both the simulated and captured datasets, we use $\lambda_{\text{eik}} = \lambda_{\text{ds}} =  0.1$

\paragraph{MonoSDF w/ mask.} MonoSDF has shown great surface reconstruction quality in few-view settings by introducing additional monocular depth and normal supervision. In particular, they leverage the pretrained Omnidata~\cite{eftekhar2021omnidata} model to estimate the depths and normals of training images. For a fair comparison, we instead estimate the depths from the measured transient using a log-matched filter~\cite{kirmani2014first}. As the estimated depths are noisy, and to provide additional supervision on the background, we used estimated masks to segment the object of interest, provide masked supervision on the depth and images~\cite{mildenhall2021nerf}.

From the estimated depth, we estimate normals by calculating the finite difference gradients of the depth point clouds. In particular, let $\tilde{d}$ be the estimated depth of a pixel in an image from a view, we use the camera parameters of that view to compute origins $\bm{o}$ and viewing directions $\bm{v}$ for each pixel. The point cloud $\bm{pc}$ for pixel $(i, j)$ can be calculated as $\bm{pc}_{ij} = \tilde{d}\bm{v} + \bm{o}$. Next, we compute gradients in the horizontal ($x$) and vertical ($y$) directions, $g_x = \frac{\bm{pc}_{i+2j}-\bm{pc}_{ij}}{2}$ and $g_y = \frac{\bm{pc}_{ij+2}-\bm{pc}_{ij}}{2}$ and compute their normal as $\bm{n} = \frac{g_x \times g_y}{||g_x \times g_y||}_2$. Lastly, for fair comparison, we adopt the NeuS \cite{wang2021neus} conversion of the SDF to density. The training loss is $\mathcal{L} = \mathcal{L}_{\text{photo}} + \lambda_{\text{eik}}\mathcal{L}_{\text{eik}} + \lambda_{\text{depth}}\mathcal{L}_{\text{depth}} + \lambda_{\text{normal}}\mathcal{L}_{\text{normal}}$, where $\lambda_{\text{eik}} = \lambda_{\text{depth}}= 0.1$ and $\lambda_{\text{normal}} = 0.05$. \\

\subsection{Hyperparameters}
For the simulated dataset, we set \\
$\{\lambda_{\text{ref}}, \lambda_{\text{eik}}, \lambda_{\text{sc}}, \lambda_{\text{weight\_var}}, \lambda_{\text{sparse}} \} = \\
\{3\times10^{-3}, 1\times10^{-5}, 7\times10^{-3}, 1\times-10^{-3}, 3\times10^{-7}\}$. The sparsity scale $\alpha$ in the sparsity regularizer is set to 100. We do \emph{not} tune these hyperparameters per scene. For the low photon experiments, since the overall signal of the transient becomes weaker, we need to enforce stricter supervision of the reflectivity loss. In particular, for photon experiments with $300, 150$ photons per pixel, we increase $\lambda_{\text{ref}} = 5\times 10^{-3}$, for the experiment with 50ppp, we increase $\lambda_{\text{ref}} = 6\times 10^{-3}$, and for the experiment with 10ppp, we set $\lambda_{\text{ref}} = 2\times 10^{-2}$.

For the captured dataset, we set $\{\lambda_{\text{ref}}, \lambda_{\text{eik}}, \lambda_{\text{sc}}, \lambda_{\text{weight\_var}}, \lambda_{\text{sparse}} \} = \\
\{7\times10^{-3}, 1\times10^{-5}, 1\times10^{-2}, 3\times-10^{-2}, 1\times10^{-4}\}$ and increase the weight of the reflectivity loss $\lambda_{ref}$ to $2\times10^{-2}$ for the experiments with $10$ ppp training transients, which improves convergence.

\subsection{Transient IoU Metric}
In the main results of the paper, we compare our rendered transients against baseline methods, using the Transient IOU metric, which was previously introduced in Flying with Photons~\cite{malik2024flying}. The metric measures the intersection over union of the predicted transient against the measured transient. Formally, let $\bm{\tau}_f$ and $\bm{\tau}$ be the predicted and measured transients, respectively. The Transient IoU is defined as:
\begin{equation*}
    \text{IoU}(\bm{\tau}_f, \bm{\tau}) =  \frac{\sum \min(\bm{\tau}_f, \bm{\tau})}{\sum \min(\bm{\tau}_f, \bm{\tau})},
\end{equation*}
where the minimum and maximum are computed element-wise and the summation is summed across all dimensions of the tensor. The IoU metric ranges from 0 to 1 and measures the overlap between the predicted and measured transients. A score of 0 indicates no overlap while a score of 1 indicates perfect overlap.

\subsection{Low Photon Count Dataset}
To create the low photon count transient dataset in simulation for each photon level $\{300, 150, 50, 10\}$, we integrate the transients along the time dimension to get images, and solve for a scale factor such that the ratio between the sum of pixels and the sum of the masked pixels of all images equals the desired photon level. After, we scale down the training transients by the scale factor and add a background level, which is set to 0.001 photons per 2850 scene photons to approximately match the captured results~\cite{malik2023transient}. We then Poisson sample the resulting transient for each photon level. \\

To create the low photon dataset for the captured dataset, we again estimate the average number of photons per occupied pixel. To synthesize a transient with a desired average number of photons per pixel, we use subsample the photon arrivals, in a process called thinning~\cite{lewis1979simulation}.

\section{Additional Results}

\subsection{Ablation Studies}
\paragraph{Regularizers.} In the following section, we show ablation studies of each of our regularizers on the \textit{Lego} scene from the simulated dataset. We find that the method does not converge without the reflectivity loss, so we omit the quantitative results for that ablation. Our results are illustrated in Table \ref{tab:ablation_loss_complete}.

In addition to the ablations shown in the main paper we also ablate the use of the Eikonal loss. The results obtained with and without the Eikonal loss have similar performance. As mentioned in the main paper, we tuned the Eikonal loss and the weight variance loss when ablating the reflectivity loss. In particular, we set $\lambda_{\text{eik}} = 0$ and $\lambda_{\text{weight\_var}} = 1 \times 10 ^ {-5}$. Without those changes, training would diverge.

\begin{table*}
    \caption{Ablation study on the proposed method in the \textit{lego} scene. We present results while omitting the depth variance regularizer, without the Eikonal loss, without the sparsity loss, without the space carving loss, and without the reflectivity loss} 
    \label{tab:ablation_loss_complete}
    \centering
    \resizebox{\textwidth}{!}{%
    \begin{tabular}{ll|ccc|ccc|ccc|ccc|ccc}
        \toprule  
        & & \multicolumn{3}{c|}{Chamfer Distance $\,\downarrow$} & \multicolumn{3}{c|}{PSNR (dB)$\,\uparrow$} & \multicolumn{3}{c|}{LPIPS$\,\downarrow$} & \multicolumn{3}{c|}{L1 (depth) $\,\downarrow$} & \multicolumn{3}{c}{Transient IOU$\,\uparrow$}\\ 
        & \textbf{Method}                             & 2 views & 3 views & 5 views & 2 views & 3 views & 5 views & 2 views & 3 views & 5 views  & 2 views & 3 views & 5 views & 2 views & 3 views & 5 views\\\midrule

        &Ours                             & \cellcolor{tabsecond}0.02 & \cellcolor{tabsecond}0.02 & \cellcolor{tabfirst}0.01 & \cellcolor{tabfirst}25.72 & \cellcolor{tabsecond}26.66 & \cellcolor{tabsecond}27.47 & \cellcolor{tabsecond}0.197 & \cellcolor{tabsecond}0.184 & \cellcolor{tabthird}0.183 & \cellcolor{tabfirst}0.008 & \cellcolor{tabfirst}0.008 & \cellcolor{tabfirst}0.007 & \cellcolor{tabfirst}0.53 & \cellcolor{tabfirst}0.58 & \cellcolor{tabsecond}0.63 \\
        &Ours w/o $\lambda_{sc}$          & \cellcolor{tabthird}0.07 & 0.04 & \cellcolor{tabthird}0.11 & \cellcolor{tabthird}25.63 & 26.32 & 26.22 & \cellcolor{tabthird}0.210 & \cellcolor{tabthird}0.193 & 0.220 & \cellcolor{tabsecond}0.009 & \cellcolor{tabfirst}0.008 & \cellcolor{tabsecond}0.008 & \cellcolor{tabfirst}0.53 & \cellcolor{tabsecond}0.57 & 0.53 \\
        &Ours w/o $\lambda_{eik}$         & \cellcolor{tabfirst}0.01 & \cellcolor{tabfirst}0.01 & \cellcolor{tabfirst}0.01 & \cellcolor{tabsecond}25.67 & \cellcolor{tabthird}26.62 & \cellcolor{tabfirst}27.80 & \cellcolor{tabfirst}0.179 & \cellcolor{tabfirst}0.153 & \cellcolor{tabfirst}0.152 & \cellcolor{tabsecond}0.009 & \cellcolor{tabfirst}0.008 & \cellcolor{tabfirst}0.007 & \cellcolor{tabsecond}0.52 & \cellcolor{tabfirst}0.58 & \cellcolor{tabfirst}0.65 \\
        &Ours w/o $\lambda_{weight\_var}$ & 0.28 & 0.37 & 1.04 & 21.58 & 22.66 & 23.81 & 0.380 & 0.280 & 0.258 & \cellcolor{tabthird}0.010 & \cellcolor{tabfirst}0.008 & \cellcolor{tabfirst}0.007 & 0.48 & 0.53 & \cellcolor{tabthird}0.60 \\
        &Ours w/o $\lambda_{sparse}$      & \cellcolor{tabsecond}0.02 & \cellcolor{tabthird}0.03 & \cellcolor{tabsecond}0.08 & 25.43 & \cellcolor{tabfirst}26.87 & \cellcolor{tabthird}26.58 & 0.218 & \cellcolor{tabthird}0.193 & 0.194 & \cellcolor{tabsecond}0.009 & \cellcolor{tabfirst}0.008 & \cellcolor{tabfirst}0.007 & \cellcolor{tabfirst}0.53 & \cellcolor{tabfirst}0.58 & \cellcolor{tabsecond}0.63\\
        &Ours w/o $\lambda_{ref}$      & 5.50   & 2.91 & 0.53 & 22.38& 24.92 & 26.22 & 0.239 & \cellcolor{tabthird}0.193 & \cellcolor{tabsecond}0.153 & 0.010 & 0.017\cellcolor{tabsecond} & \cellcolor{tabfirst}0.007 & \cellcolor{tabthird}0.49 & \cellcolor{tabthird}0.55 & \cellcolor{tabfirst}0.65\\
        \bottomrule
    \end{tabular}}
\end{table*}

\paragraph{Argmax Depth vs. Convention NeRF Depth.}
To avoid multiple depth values skewing the expected ray termination, TransientNeRF \cite{malik2023transient} proposed to calculate the depth as the argmax of the weights along a ray:
\begin{equation*}
    d_\text{ours} = \argmax_{t_i}{w}_i.
\end{equation*}
This is in contrast to the conventional depth calculation proposed in NeRF:
\begin{equation*}
    d_\text{nerf} = \sum_{i}{w}_i\frac{t_i + t_{i-1}}{2}.
\end{equation*}
We compare using each formulation in our weight variance regularizer in table \ref{tab:argmax_depth}.

\begin{table*}
    \caption{Ablation study comparing the effect of using our argmax depth formulation in the weight variance loss versus that of using NeRF's default depth calculation in the \textit{lego} scene.} 
    \label{tab:argmax_depth}
    \centering
    \resizebox{\textwidth}{!}{%
    \begin{tabular}{ll|ccc|ccc|ccc|ccc|ccc}
        \toprule  
        & & \multicolumn{3}{c|}{Chamfer Distance $\,\downarrow$} & \multicolumn{3}{c|}{PSNR (dB)$\,\uparrow$} & \multicolumn{3}{c|}{LPIPS$\,\downarrow$} & \multicolumn{3}{c|}{L1 (depth) $\,\downarrow$} & \multicolumn{3}{c}{Transient IOU$\,\uparrow$}\\ 
        & \textbf{Method}                             & 2 views & 3 views & 5 views & 2 views & 3 views & 5 views & 2 views & 3 views & 5 views  & 2 views & 3 views & 5 views & 2 views & 3 views & 5 views\\\midrule

        &$\mathcal{L}_\text{weight\_var}$ with $d_\text{ours}$          & \cellcolor{tabfirst}0.02 & \cellcolor{tabfirst}0.02 & \cellcolor{tabfirst}0.01 & \cellcolor{tabfirst}25.72 & \cellcolor{tabfirst}26.66 & \cellcolor{tabfirst}27.47 & \cellcolor{tabfirst}0.197 & 0.184 &0.183 & \cellcolor{tabfirst}0.008 & \cellcolor{tabfirst}0.008 & \cellcolor{tabfirst}0.007 & \cellcolor{tabfirst}0.53 & \cellcolor{tabfirst}0.58 & \cellcolor{tabfirst}0.63 \\
        &$\mathcal{L}_\text{weight\_var}$ with $d_\text{nerf}$       & 6.29 & 3.31 & 0.03 & 17.13 & 24.73 & 26.63 & 0.307 & \cellcolor{tabfirst}0.170 & \cellcolor{tabfirst}0.161 & 0.24 & 0.01 & 0.008 & 0.19 & 0.55 &0.61\\
        \bottomrule
    \end{tabular}}
\end{table*}

\subsection{Simulated Results}
Tables~\ref{tab:simulated-cd}, \ref{tab:simulated-psnr}, \ref{tab:simulated-lpips}, and \ref{tab:simulated-ssim} provide a breakdown of the quantitative results in simulation across all scenes and photon levels. The results per scene are largely consistent with the general trends, averaged across scenes. 

\begin{figure*}[t]
    \centering
    \includegraphics[width=0.9\textwidth]{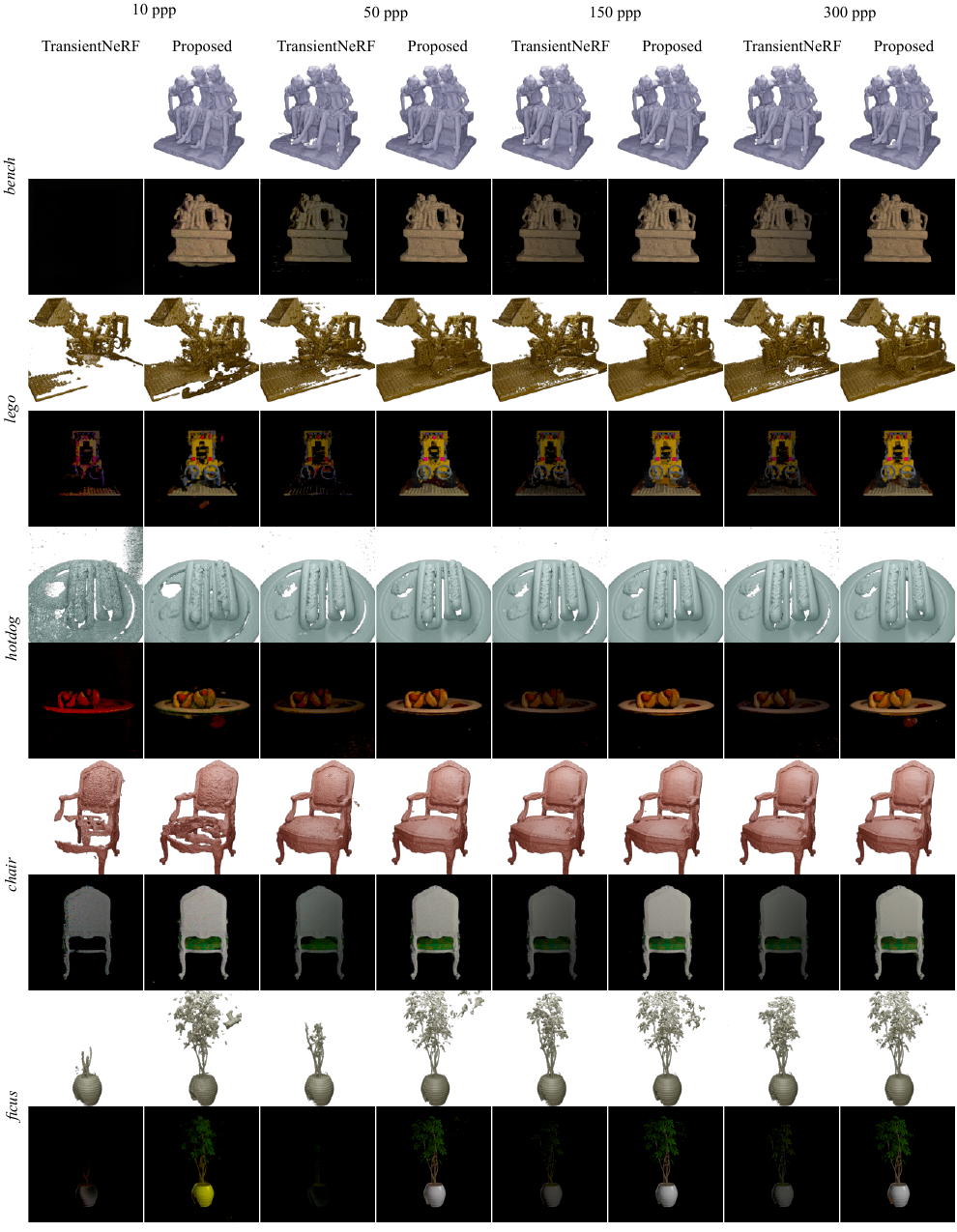}
    \caption{Rendered meshes and images for the low photon count experiments on the simulated dataset for 2 views.}
    \label{fig:sim2lp}
\end{figure*}

\begin{figure*}[t]
    \centering
    \includegraphics[width=0.9\textwidth]{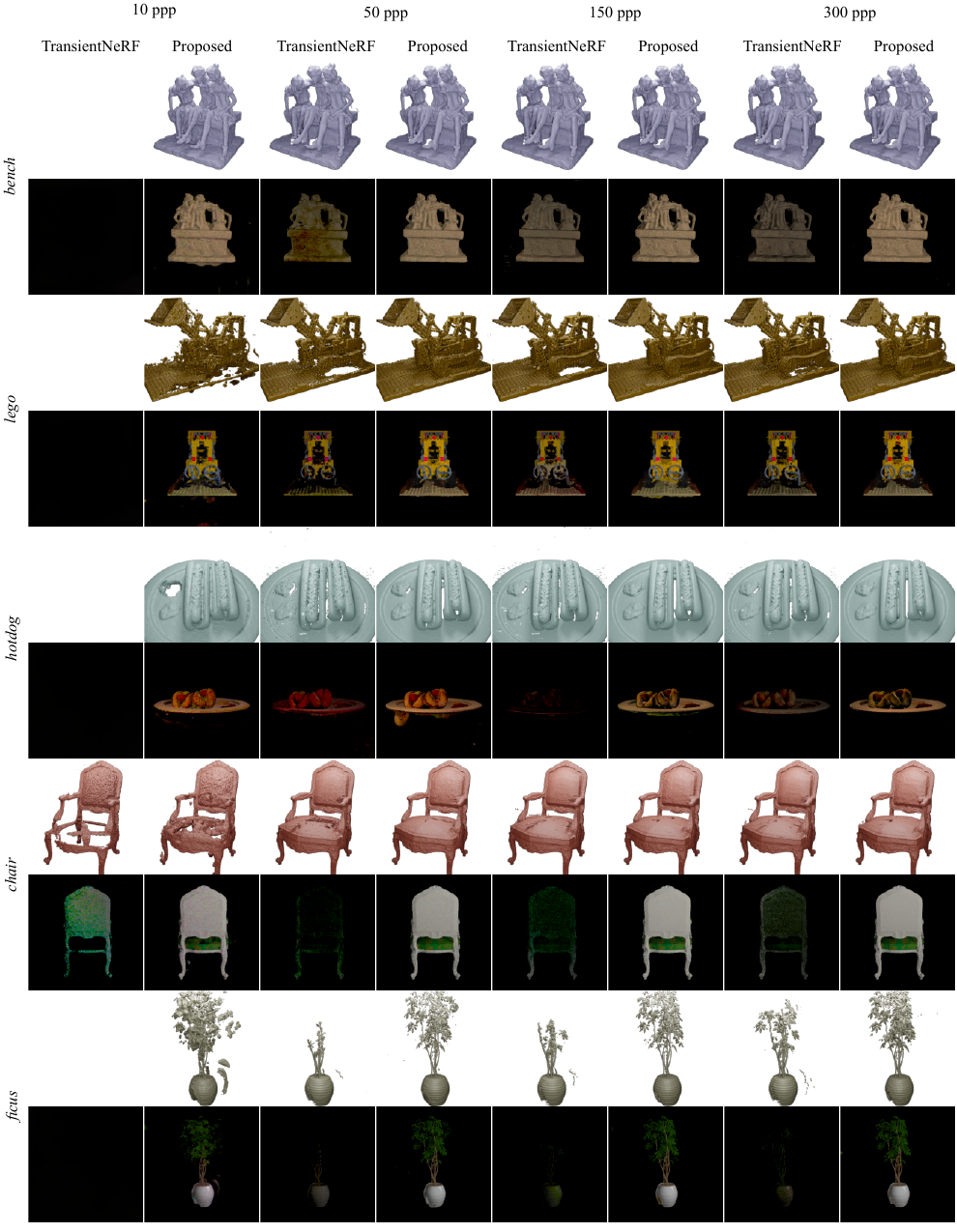}
    \caption{Rendered meshes and images for the low photon count experiments on the simulated dataset for 3 views.}
    \label{fig:sim3lp}
\end{figure*}

\begin{figure*}[t]
    \centering
    \includegraphics[width=0.9\textwidth]{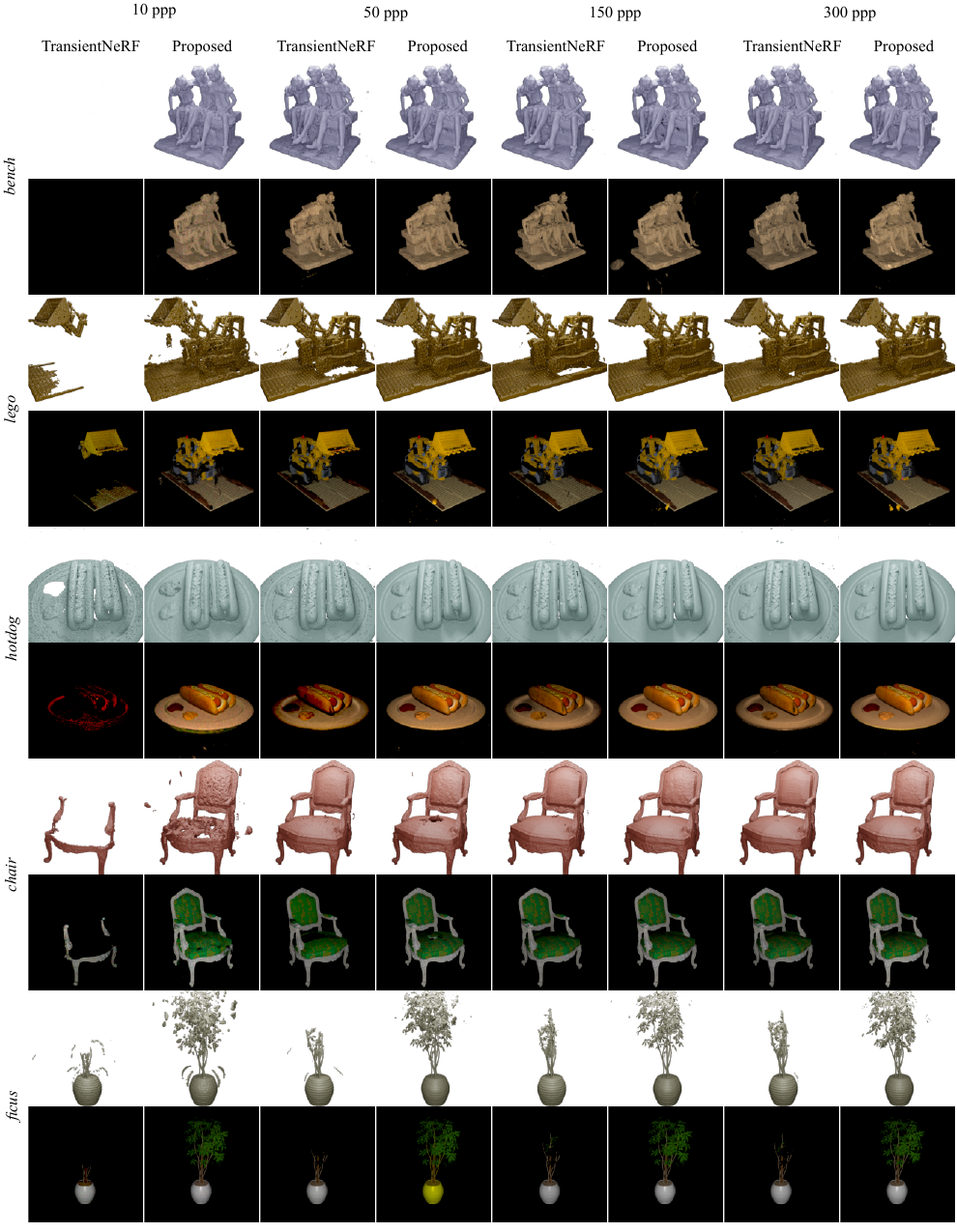}
    \caption{Rendered meshes and images for the low photon count experiments on the simulated dataset for 5 views.}
    \label{fig:sim5lp}
\end{figure*}

In Figures~\ref{fig:sim2lp}, \ref{fig:sim3lp}, and \ref{fig:sim5lp} we provide additional qualitative results across varying photon levels. As can be seen, TransientNeRF~\cite{malik2023transient} oftentimes fails to reconstruct the object in the case of $10$ photons per pixel---we indicate such cases by omitting the mesh visualization in the figure. We attribute this failure to the lack of a reflectivity loss, which we observe helps our method converge robustly even at the lowest photon levels. 

\subsection{Captured Results}
Tables ~\ref{tab:captured-psnr}, \ref{tab:captured-lpips}, \ref{tab:captured-ssim}, \ref{tab:captured-l1depth} provide a breakdown of the quantitative results in the captured dataset across all scenes and photon levels. The results are largely consistent with general trends, averaged acorss scenes.

\begin{figure*}[t]
    \centering
    \includegraphics[width=0.75\textwidth]{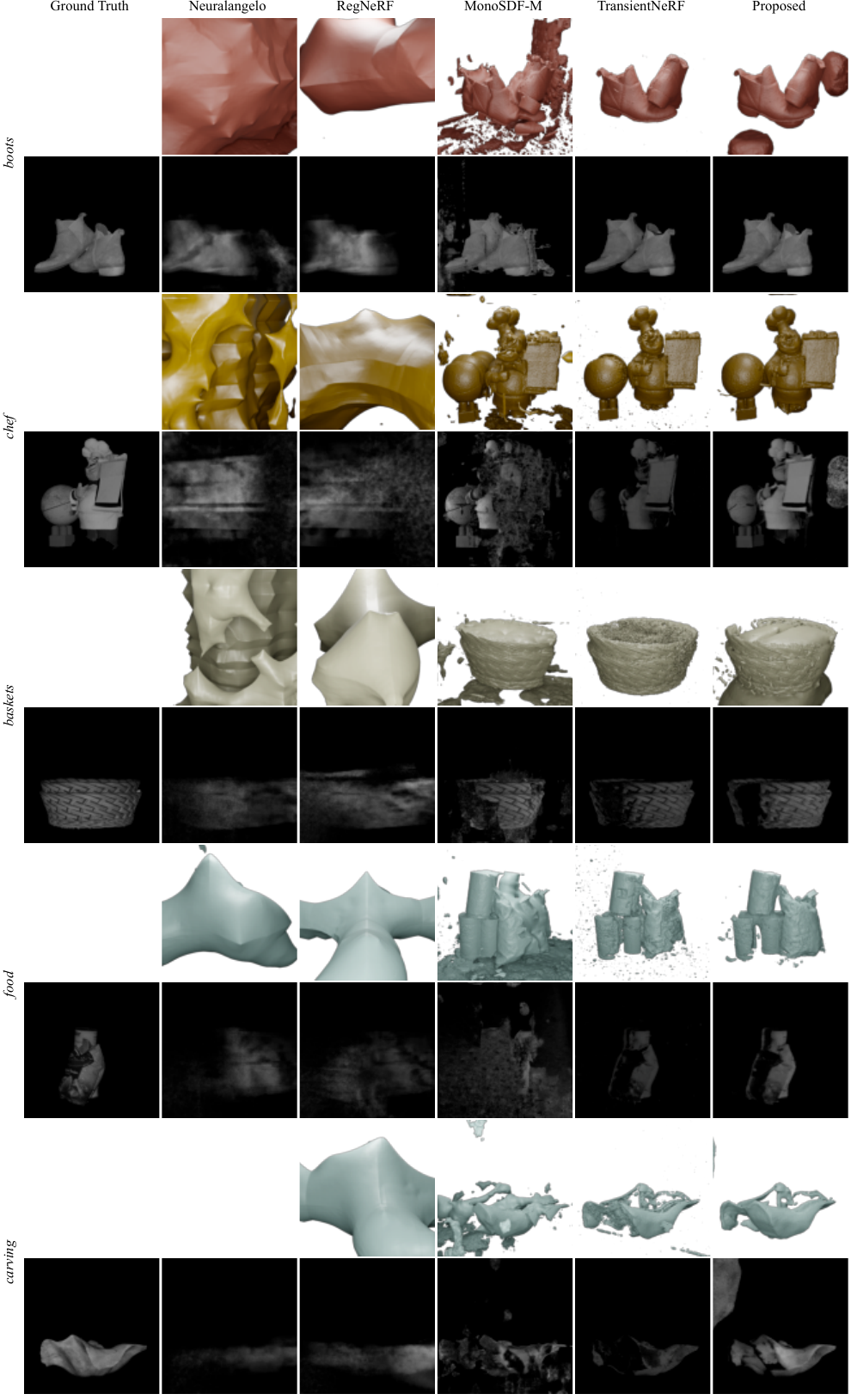}
    \caption{Rendered meshes and images on the captured dataset for 2 views.}
    \label{fig:cap2}
\end{figure*}

\begin{figure*}[t]
    \centering
    \includegraphics[width=0.75\textwidth]{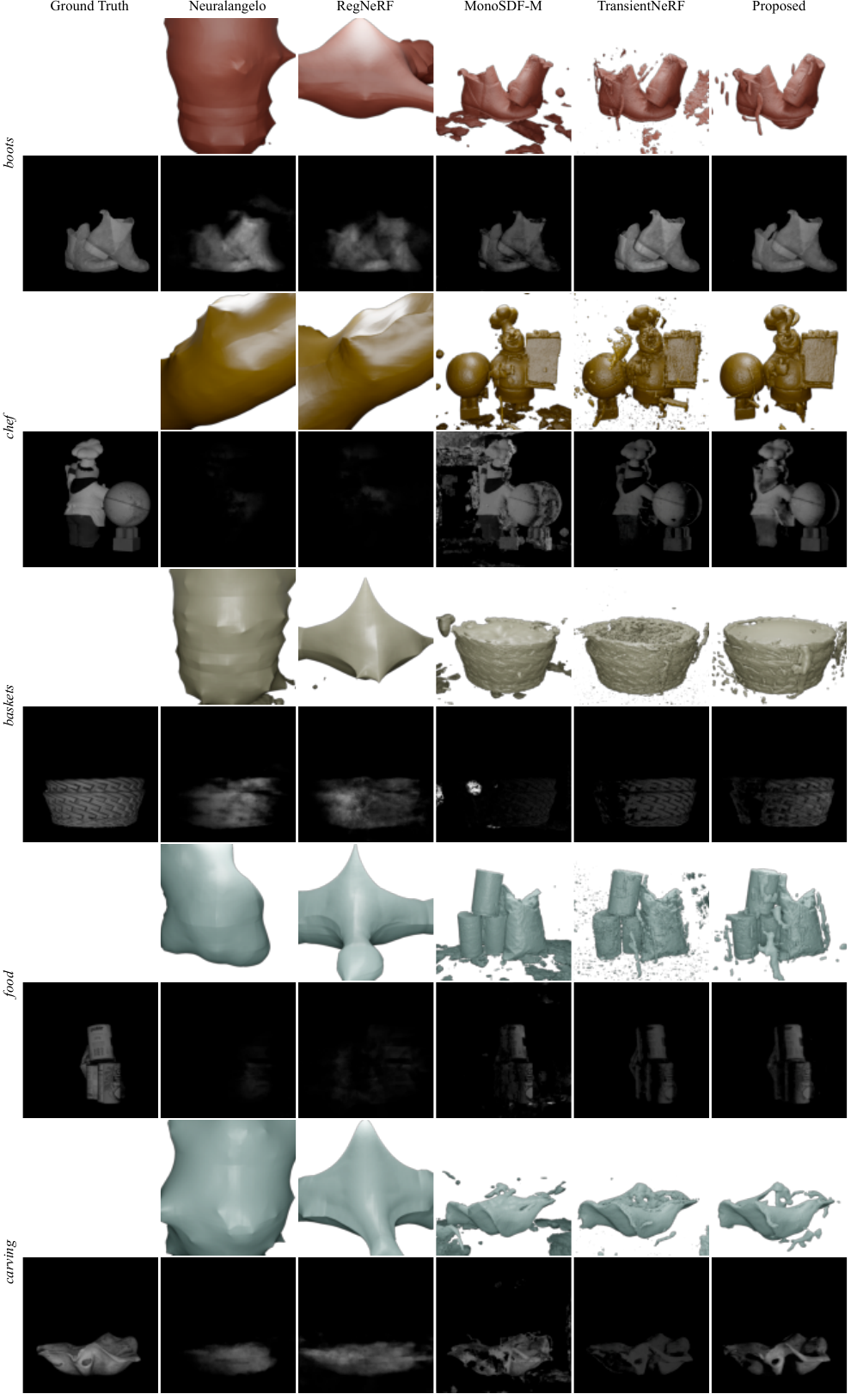}
    \caption{Rendered meshes and images on the captured dataset for 3 views.}
    \label{fig:cap3}
\end{figure*}

\begin{figure*}[t]
    \centering
    \includegraphics[width=0.75\textwidth]{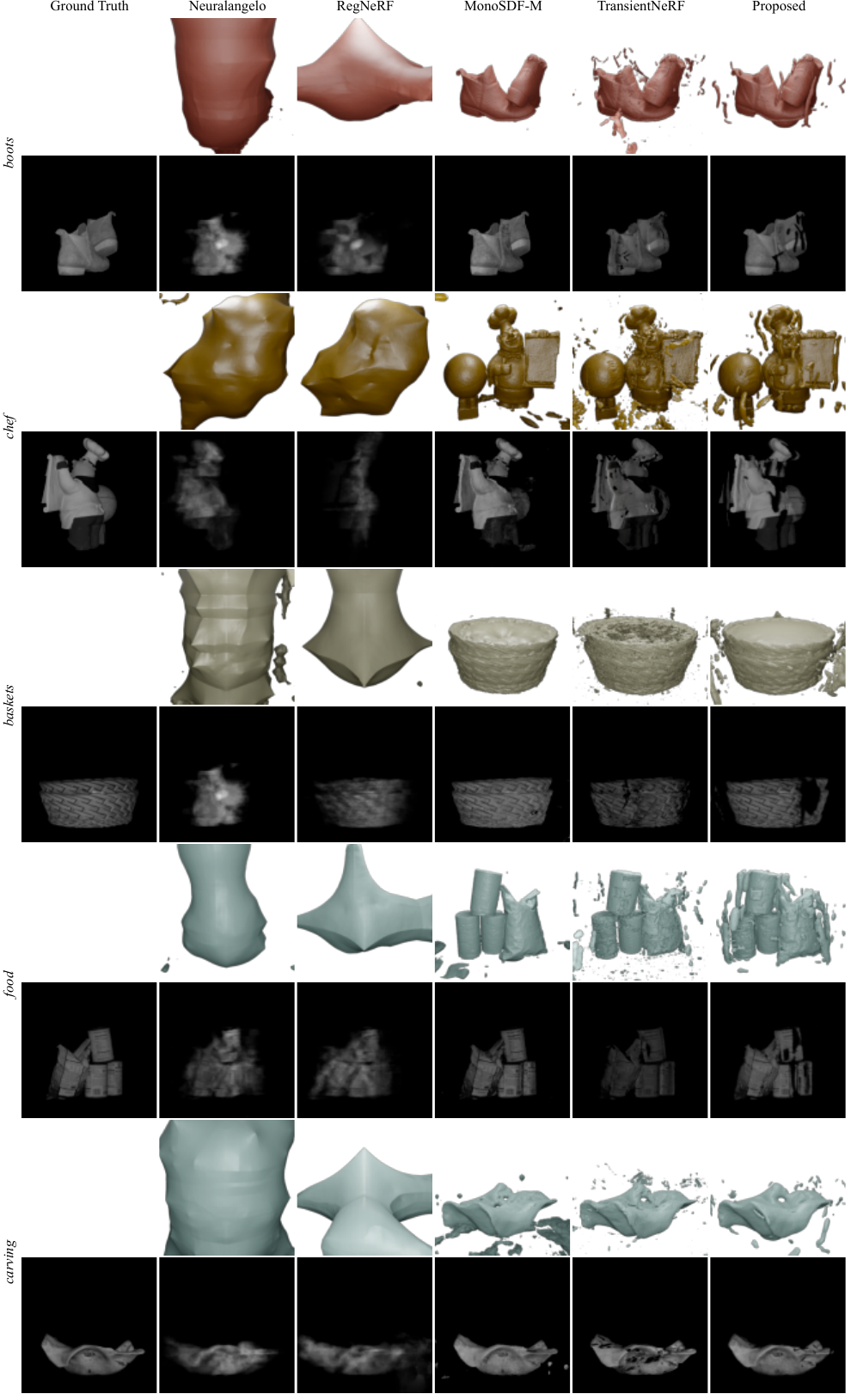}
    \caption{Rendered meshes and images on the captured dataset for 5 views.}
    \label{fig:cap5}
\end{figure*}

In Figures~\ref{fig:cap2}, \ref{fig:cap3}, and \ref{fig:cap5} we provide additional qualitative results for scenes with an average of $1500$ photons per pixel. 

\begin{figure*}[t]
    \centering
    \includegraphics[width=0.9\textwidth]{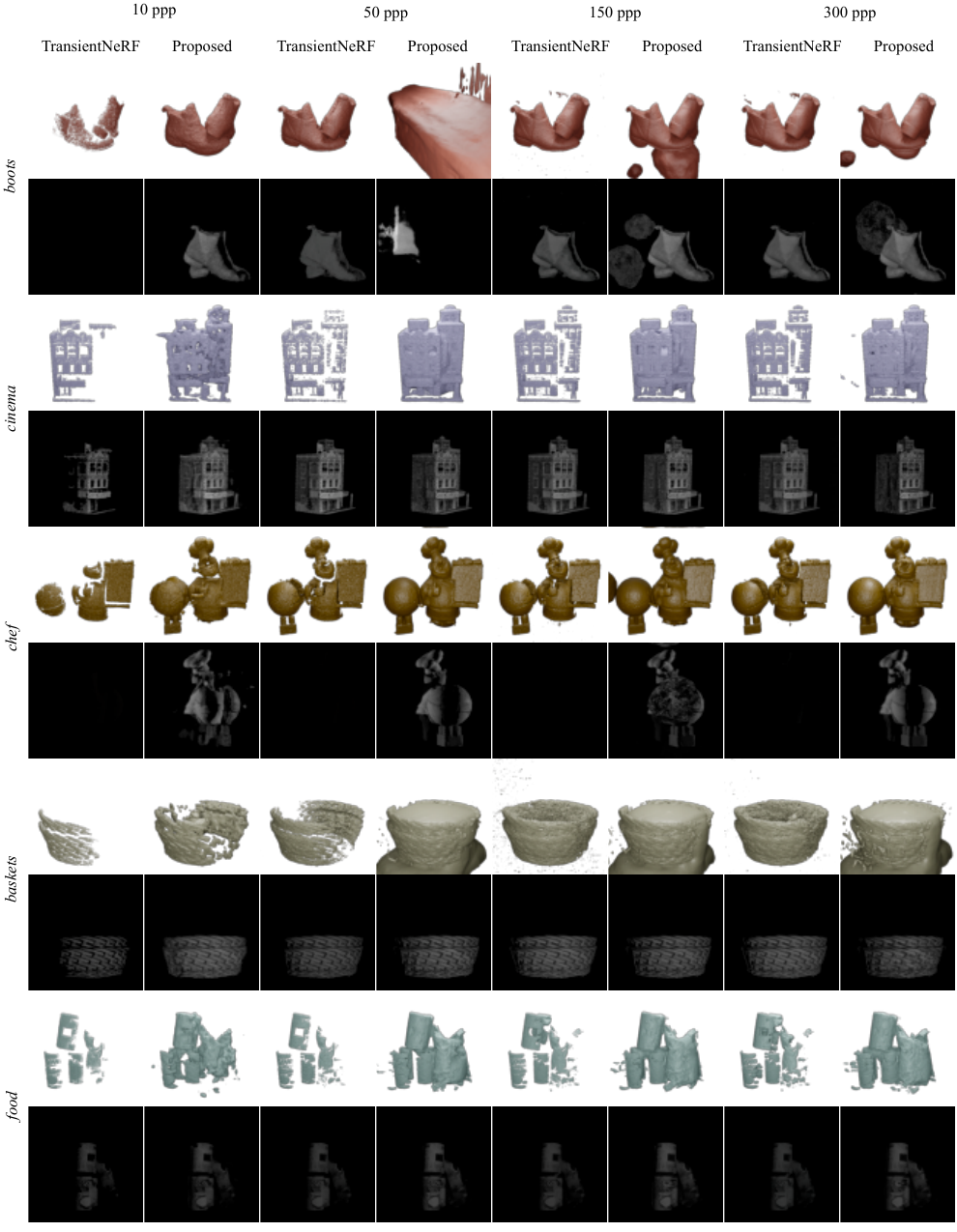}
    \caption{Rendered meshes and images for the low photon count experiments on the captured dataset for 2 views.}
    \label{fig:cap2lp}
\end{figure*}

\begin{figure*}[t]
    \centering
    \includegraphics[width=0.9\textwidth]{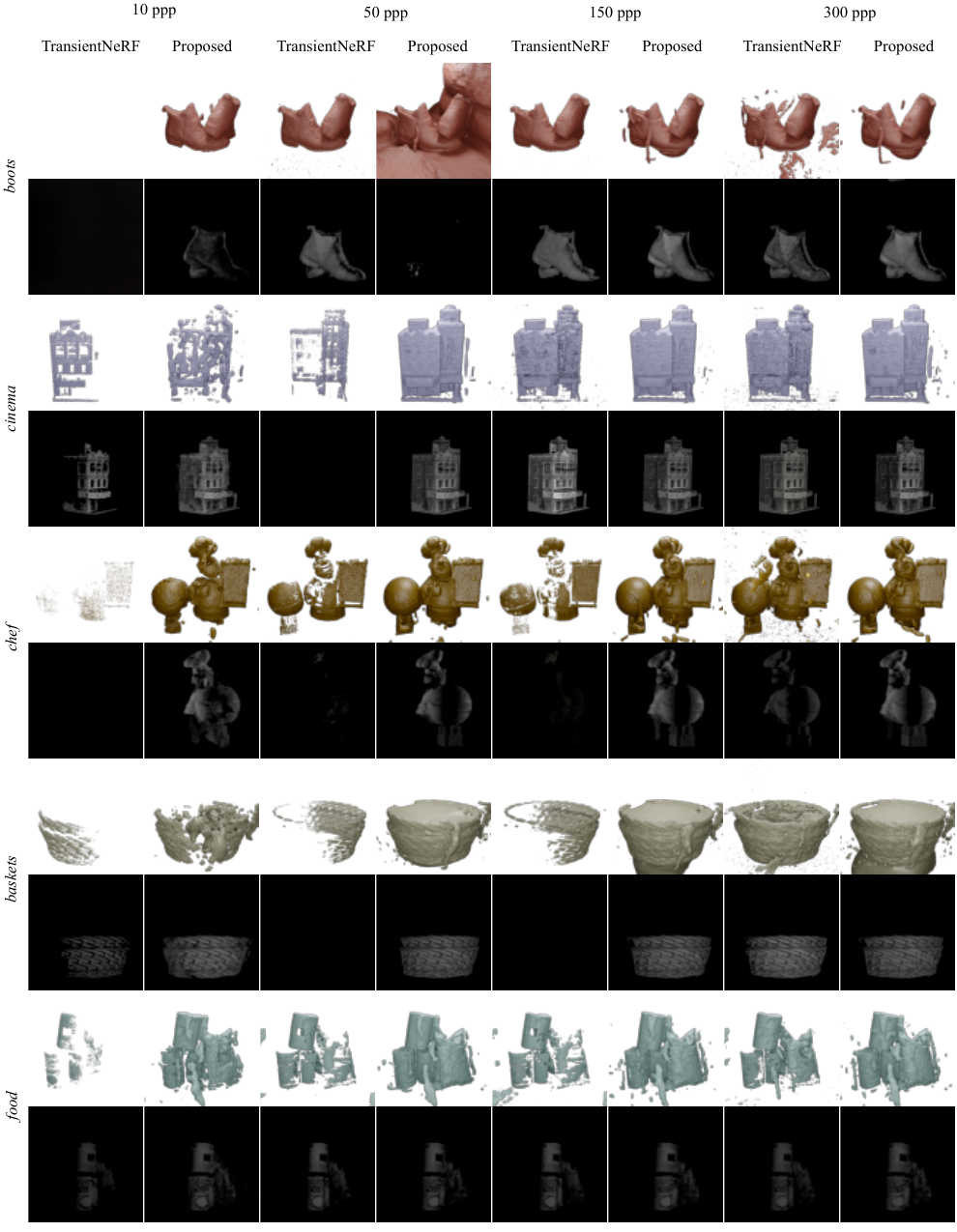}
    \caption{Rendered meshes and images for the low photon count experiments on the captured dataset for 3 views.}
    \label{fig:cap3lp}
\end{figure*}

\begin{figure*}[t]
    \centering
    \includegraphics[width=0.9\textwidth]{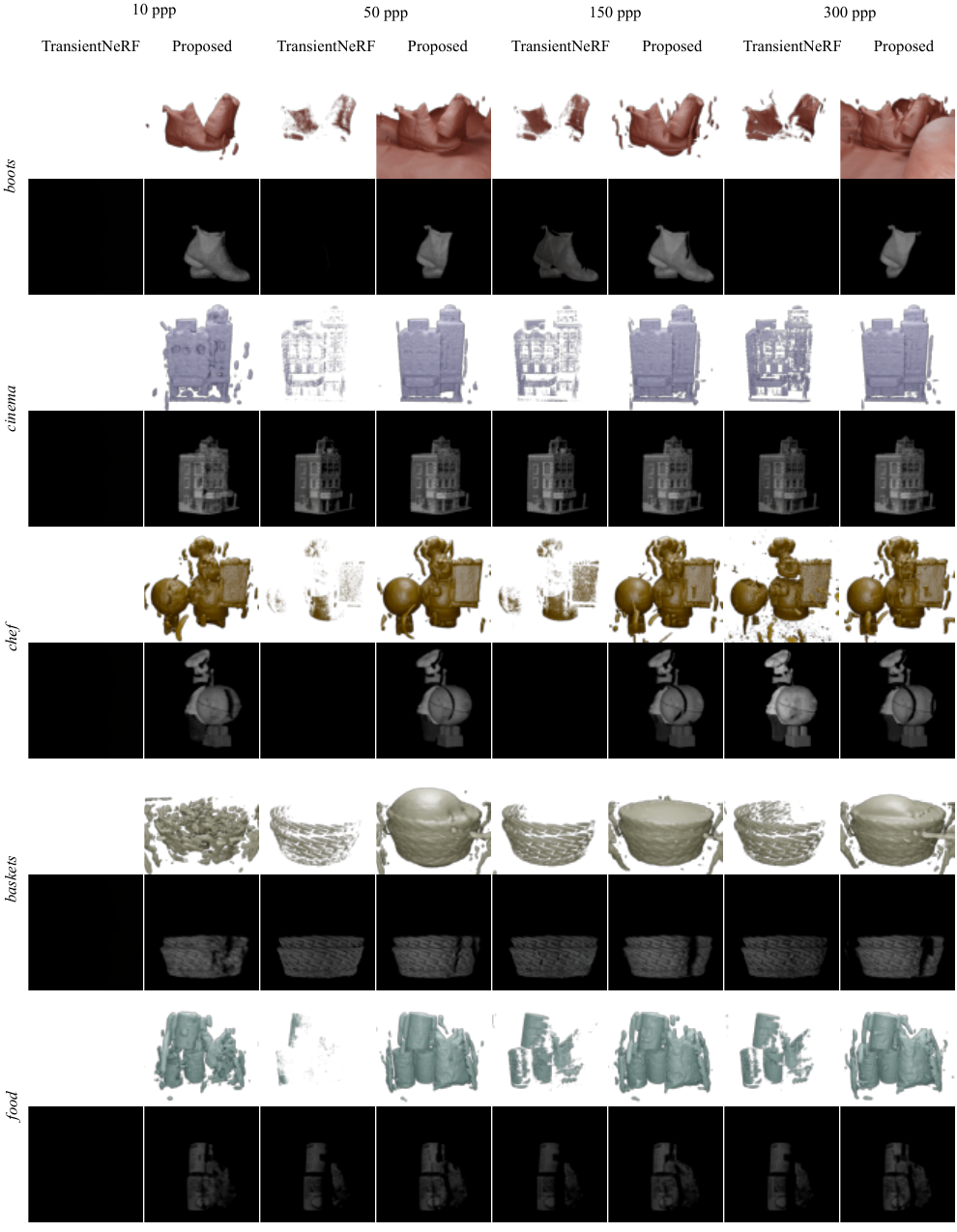}
    \caption{Rendered meshes and images for the low photon count experiments on the captured dataset for 5 views.}
    \label{fig:cap5lp}
\end{figure*}

In Figures~\ref{fig:cap2lp}, \ref{fig:cap3lp}, and \ref{fig:cap5lp} we provide additional qualitative results across varying photon levels. As can be seen, TransientNeRF~\cite{malik2023transient} oftentimes fails to reconstruct the object for scenes with an average of $10$ photons per pixel---we again indicate such cases by omitting the mesh visualization in the figure.

\subsection{Low-Photon Captured Results Without Masking}
In Table \ref{tab:captured_not_masked}, we show the results of our method and that of TransientNeRF in the low-photon regime when we are not applying ground-truth segmentation masks to filter out background noise. Compared with Table 2 in the main paper, our method drastically improves when supervised on segmented transients.

\subsection{Double Bounce}
\begin{figure*}[t]
    \centering
    \includegraphics[width=0.9\textwidth]{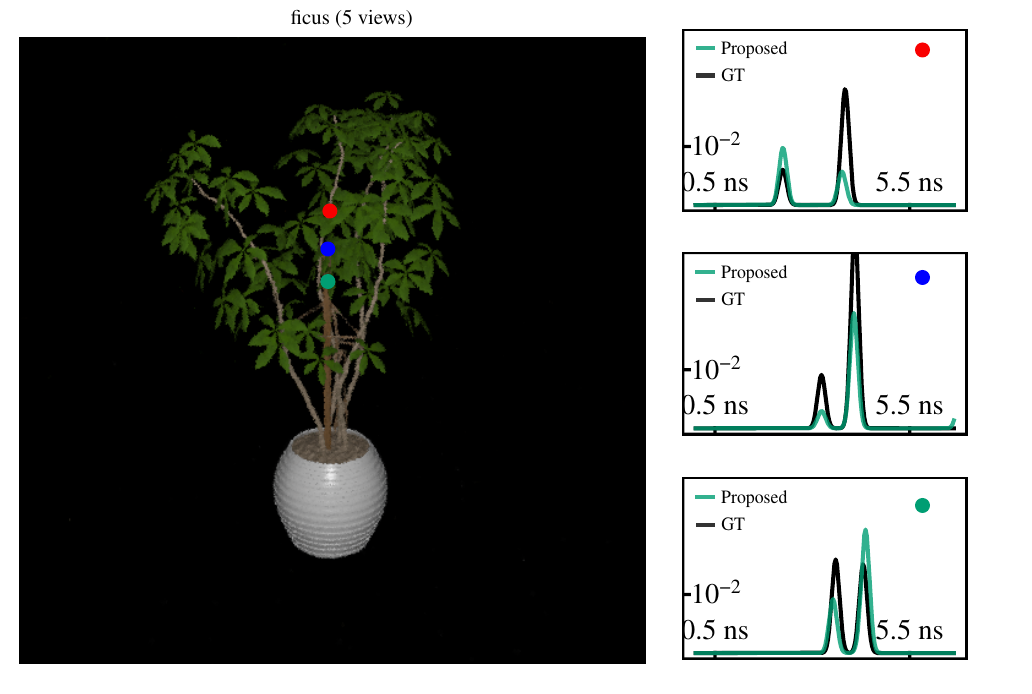}
    \caption{Predicted vs ground-truth double-bounce transients plotted versus time in (nanoseconds) for the selected pixels (red, blue, and green) for the \textit{ficus} scene, trained on five views.}
    \label{fig:double_bounce}
\end{figure*}

In Figure \ref{fig:double_bounce}, we show on the left a rendered image of the \textit{ficus} scene trained on 5 views and three selected pixel coordinates (right, blue, green) and their corresponding double-bounce transients on the right. From the transient figures, our method is fairly robust to double bounce returns, as the predicted transients match the ground-truth transients at those locations.

\subsection{Failure Case}
\begin{figure*}[t]
    \centering
    \includegraphics[width=1\textwidth]{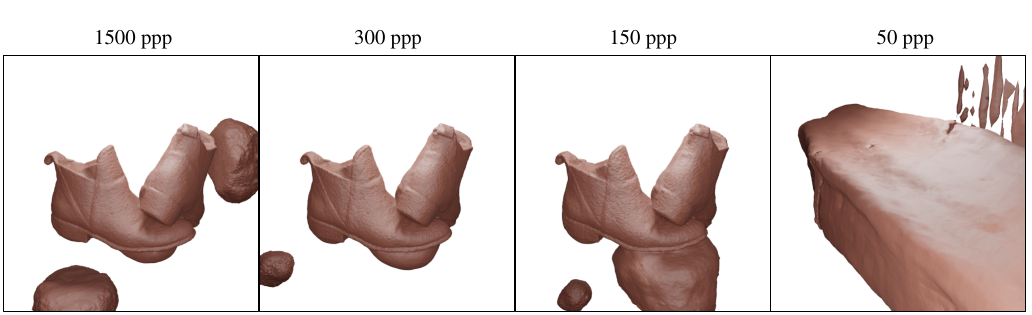}
    \caption{Failure case for our method. Rendered mesh for 1500 ppp, 300 ppp, 150 ppp, and 50 ppp where blob-like artifacts are surrounding the mesh.}
    \label{fig:failure_case}
\end{figure*}

In Figure \ref{fig:failure_case}, we show the extracted mesh from our optimized representation for the \textit{boots} scene trained from two views across multiple photon levels (1500 ppp, 300 ppp, 150 ppp, 50 ppp). Multiple blob-like artifacts surrounding the mesh are present. We attribute this failure case to the ill-posed nature of the problem due to the sparse number of viewpoints.

\begin{table*}
    \caption{Captured results assessing image quality and depth accuracy without applying segmentation masks. The left column indicates average photons per pixel.}
    \label{tab:captured_not_masked}
    \centering
    \resizebox{\textwidth}{!}{%
    \begin{tabular}{ll|ccc|ccc|ccc|ccc}
        \toprule  
        & & \multicolumn{3}{c|}{PSNR (dB) $\,\uparrow$} & \multicolumn{3}{c|}{LPIPS $\,\downarrow$} & \multicolumn{3}{c|}{L1 (depth) $\,\downarrow$} & \multicolumn{3}{c}{Transient IOU $\,\uparrow$} \\ 
        & \textbf{Method} & 2 views & 3 views & 5 views & 2 views & 3 views & 5 views & 2 views & 3 views & 5 views & 2 views & 3 views & 5 views\\
        \midrule

        \parbox[t]{6mm}{\multirow{2}{*}{\rotatebox[origin=c]{90}{\textit{300}}}} 
        & TransientNeRF & \cellcolor{red}19.83 & \cellcolor{red}21.39 & 21.61 & \cellcolor{red}0.310 & \cellcolor{red}0.227 & \cellcolor{red}0.187 &\cellcolor{red} 0.013 & \cellcolor{red}0.001 & 0.012 & 0.25 & 0.36 & 0.39 \\
        & Proposed & 19.25 & 21.18 & \cellcolor{red}23.97 & 0.372 & 0.321 & 0.318 & 0.024 & 0.019 & \cellcolor{red}0.011 & \cellcolor{red}0.34 & \cellcolor{red}0.41 & \cellcolor{red}0.57 \\
        \midrule
        \parbox[t]{6mm}{\multirow{2}{*}{\rotatebox[origin=c]{90}{\textit{150}}}} 
        & TransientNeRF \hspace{1em} & 19.25 & \cellcolor{red}21.30 & 19.85 & \cellcolor{red}0.282 & \cellcolor{red}0.217 & \cellcolor{red}0.196 & \cellcolor{red}0.016 &\cellcolor{red} 0.012 & 0.017 & 0.22 & 0.35 & 0.28 \\
        & Proposed \hspace{1em} & \cellcolor{red}19.30 & 20.79 & \cellcolor{red}23.49 & 0.387 & 0.358 & 0.298 & 0.025 & 0.020 & \cellcolor{red}0.011 & \cellcolor{red}0.32 & \cellcolor{red}0.39 & \cellcolor{red}0.55 \\                
        \midrule
        \parbox[t]{6mm}{\multirow{2}{*}{\rotatebox[origin=c]{90}{\textit{50}}}} 
        & TransientNeRF \hspace{1em} & 18.98 & \cellcolor{red}19.95 & 18.42 & 0.284 & \cellcolor{red}0.207 & \cellcolor{red}0.205 & 0.019 & \cellcolor{red}0.020 &\cellcolor{red} 0.021 & 0.21 & 0.28 & 0.19 \\
        & Proposed \hspace{1em} & \cellcolor{red}21.54 & 19.47 & \cellcolor{red}23.55 & \cellcolor{red}0.238 & 0.246 & 0.227 & \cellcolor{red}0.006 & 0.035 & 0.077 & \cellcolor{red}0.42 & \cellcolor{red}0.30 & \cellcolor{red}0.54 \\
        \midrule
        \parbox[t]{6mm}{\multirow{2}{*}{\rotatebox[origin=c]{90}{\textit{10}}}} 
        & TransientNeRF \hspace{1em} & \cellcolor{red}16.19 & 15.87 & 15.70 & \cellcolor{red}0.257 & 0.264 & 0.274 & \cellcolor{red}0.039 & 0.056 & 0.080 & 0.08 & 0.05 & 0.03 \\
        & Proposed \hspace{1em} & 15.03 & \cellcolor{red}20.62 & \cellcolor{red}24.02 & 0.379 & \cellcolor{red}0.220 & \cellcolor{red}0.168 & 0.063 & \cellcolor{red}0.025 & \cellcolor{red}0.015 & \cellcolor{red}0.13 & \cellcolor{red}0.34 & \cellcolor{red}0.53 \\
        \bottomrule
    \end{tabular}}
\end{table*}

\begin{table*}
    \caption{Breakdown of Chamfer Distances across all 5 simulated scenes}
    \label{tab:simulated-cd}
    \centering
    \resizebox{\textwidth}{!}{%
    \begin{tabular}{ll|ccc|ccc|ccc|ccc|ccc}
        \toprule  
         & & \multicolumn{3}{c|}{Neuralangelo~\cite{li2023neuralangelo}} & \multicolumn{3}{c|}{RegNeRF~\cite{niemeyer2022regnerf}} & \multicolumn{3}{c|}{MonoSDF-M~\cite{yu2022monosdf}} & \multicolumn{3}{c|}{TransientNeRF~\cite{malik2023transient}} & \multicolumn{3}{c}{Ours} \\
        & \textbf{Scene} & 2 views & 3 views & 5 views & 2 views & 3 views & 5 views & 2 views & 3 views & 5 views & 2 views & 3 views & 5 views & 2 views & 3 views & 5 views \\\midrule
        \parbox[t]{6mm}{\multirow{5}{*}{\rotatebox[origin=c]{90}{\textit{6000}}}} 
        &Lego    & 4.03 & 3.02 & 3.31 & 2.95 & 2.12 & 2.76 & 1.50 & 0.28 & 0.07 & 0.05 & 0.08 & 0.05 & 0.02 & 0.02 & 0.01 \\
        &Chair   & 3.59 & 6.63 & 4.98 & 3.46 & 3.23 & 5.73 & 0.34 & 0.04 & 0.02 & 0.13 & 0.14 & 0.08 & 0.04 & 0.06 & 0.04 \\
        &Ficus   & 1.85 & 1.49 & 1.76 & 1.92 & 1.57 & 2.34 & 5.64 & 3.63 & 0.48 & 0.10 & 0.05 & 0.09 & 0.02 & 0.08 & 0.03 \\
        &Hotdog  & 7.71 & 3.54 & 2.92 & 12.59 & 3.90 & 1.94 & 1.48 & 1.02 & 0.29 & 0.66 & 1.45 & 1.06 & 0.22 & 0.06 & 0.03 \\
        &Bench   & 5.60 & 25.27 & 22.89 & 3.78 & 6.57 & 9.13 & 9.92 & 3.37 & 0.04 & 0.44 & 0.50 & 0.27 & 0.03 & 0.10 & 0.12 \\
        \midrule
        &Average & 4.56 & 7.99 & 7.17 & 4.94 & 3.48 & 4.38 & 3.78 & 1.67 & 0.18 & 0.28 & 0.44 & 0.31 & 0.07 & 0.06 & 0.05 \\
        \midrule
        \parbox[t]{6mm}{\multirow{5}{*}{\rotatebox[origin=c]{90}{\textit{300}}}} 
        &Lego    & - & - & - & - & - & – & - & - & - & 0.12 & 0.06 & 0.05 & 0.01 & 0.02 & 0.02 \\
        &Chair   & - & - & - & - & - & - & - & - & - & 0.17 & 0.17 & 0.10 & 0.06 & 0.03 & 0.07 \\
        &Ficus   & - & - & - & - & - & - & - & - & - & 0.11 & 0.46 & 0.57 & 0.05 & 0.12 & 0.08 \\
        &Hotdog  & - & - & - & - & - & - & - & - & - & 0.28 & 0.49 & 0.50 & 0.11 & 0.04 & 0.02 \\
        &Bench   & - & - & - & - & - & - & - & - & - & 0.60 & 0.47 & 0.28 & 0.11 & 0.10 & 0.12 \\
        \midrule
        &Average & - & - & - & - & - & - & - & - & - & 0.26 & 0.33 & 0.30 & 0.07 & 0.06 & 0.06 \\

        \midrule
        \parbox[t]{6mm}{\multirow{5}{*}{\rotatebox[origin=c]{90}{\textit{150}}}} 
        &Lego    & - & - & - & - & - & – & - & - & - & 0.09 & 0.15 & 0.04 & 0.02 & 0.05 & 0.04 \\
        &Chair   & - & - & - & - & - & - & - & - & - & 0.26 & 0.15 & 0.09 & 0.09 & 0.05 & 0.03 \\
        &Ficus   & - & - & - & - & - & - & - & - & - & 0.11 & 1.14 & 0.64 & 0.20 & 0.06 & 0.04 \\
        &Hotdog  & - & - & - & - & - & - & - & - & - & 0.20& 1.18 & 0.25 & 0.05 & 0.05 & 0.04 \\
        &Bench   & - & - & - & - & - & - & - & - & - & 0.60& 0.62 & 0.31 & 0.11 & 0.13 & 1.32 \\
        \midrule
        &Average & - & - & - & - & - & - & - & - & - & 0.25 & 0.65 & 0.27 & 0.09 & 0.07 & 0.29 \\
        \midrule
        \parbox[t]{6mm}{\multirow{5}{*}{\rotatebox[origin=c]{90}{\textit{10}}}} 
        &Lego    & - & - & - & - & - & – & - & - & - & 0.87 & 0.35 & 8.48 & 0.41 & 0.35 & 0.08 \\
        &Chair   & - & - & - & - & - & - & - & - & - & 0.97 & 0.30 & 5.56 & 0.09 & 0.13 & 0.25 \\
        &Ficus   & - & - & - & - & - & - & - & - & - & 3.91 & 0.36 & 3.20 & 0.18 & 0.36 & 0.21 \\
        &Hotdog  & - & - & - & - & - & - & - & - & - & 8.38 & 0.43 & 0.26 & 0.25 & 0.43 & 0.28 \\
        &Bench   & - & - & - & - & - & - & - & - & - & 0.15 & 0.12 & 10.02 & 0.15 & 0.12 & 0.07 \\
        \midrule
        &Average & - & - & - & - & - & - & - & - & - & 2.86 & 0.31 & 5.50 & 0.22 & 0.28 & 0.18 \\
        \bottomrule
    \end{tabular}}
\end{table*}

\begin{table*}
    \caption{Breakdown of PSNR across all 5 simulated scenes}
    \label{tab:simulated-psnr}
    \centering
    \resizebox{\textwidth}{!}{%
    \begin{tabular}{ll|ccc|ccc|ccc|ccc|ccc}
        \toprule  
         & & \multicolumn{3}{c|}{Neuralangelo~\cite{li2023neuralangelo}} & \multicolumn{3}{c|}{RegNeRF~\cite{niemeyer2022regnerf}} & \multicolumn{3}{c|}{MonoSDF-M~\cite{yu2022monosdf}} & \multicolumn{3}{c|}{TransientNeRF~\cite{malik2023transient}} & \multicolumn{3}{c}{Ours} \\
        & \textbf{Scene} & 2 views & 3 views & 5 views & 2 views & 3 views & 5 views & 2 views & 3 views & 5 views & 2 views & 3 views & 5 views & 2 views & 3 views & 5 views \\\midrule
        \parbox[t]{6mm}{\multirow{5}{*}{\rotatebox[origin=c]{90}{\textit{6000}}}} 
        &Lego    & 17.72 & 19.17 & 18.33 & 17.91 & 19.09 & 18.97 & 17.46 & 21.93 & 22.45 & 20.64 & 23.63 & 25.81 & 25.72 & 26.66 & 27.47 \\
        &Chair   & 17.29 & 16.92 & 24.54 & 14.87 & 17.04 & 24.64 & 21.38 & 21.44 & 30.86 & 20.75 & 21.99 & 34.48 & 25.70 & 26.98 & 31.14 \\
        &Ficus   & 23.26 & 23.63 & 25.82 & 22.97 & 23.41 & 25.32 & 23.83 & 23.75 & 26.22 & 24.57 & 26.10 & 27.70 & 28.14 & 28.35 & 32.52 \\
        &Hotdog  & 15.87 & 19.85 & 20.41 & 17.25 & 20.32 & 20.34 & 16.84 & 16.89 & 18.95 & 20.74 & 22.64 & 32.36 & 22.72 & 26.20 & 28.33 \\
        &Bench   & 18.72 & 17.65 & 21.04 & 18.58 & 19.57 & 19.08 & 19.30 & 19.68 & 24.76 & 20.20 & 23.06 & 21.57 & 24.10 & 25.24 & 27.99 \\
        \midrule
        &Average & 18.57 & 19.44 & 22.03 & 18.32 & 19.89 & 21.67 & 19.76 & 20.74 & 24.65 & 21.38 & 23.48 & 28.38 & 25.28 & 26.69 & 29.49 \\
        \midrule

        \parbox[t]{6mm}{\multirow{5}{*}{\rotatebox[origin=c]{90}{\textit{300}}}} 
        &Lego    & - & - & - & - & - & – & - & - & - & 20.42 & 23.97 & 25.04 & 26.04 & 26.79 & 26.75 \\
        &Chair   & - & - & - & - & - & - & - & - & - & 20.78 & 21.54 & 33.45 & 25.98 & 26.96 & 30.89 \\
        &Ficus   & - & - & - & - & - & - & - & - & - & 23.16 & 22.87 & 24.14 & 27.58 & 28.99 & 31.64 \\
        &Hotdog  & - & - & - & - & - & - & - & - & - & 21.55 & 22.64 & 21.89 & 24.42 & 25.36 & 26.70 \\
        &Bench   & - & - & - & - & - & - & - & - & - & 20.79 & 22.95 & 24.87 & 24.47 & 25.15 & 26.51 \\
        \midrule
        &Average & - & - & - & - & - & - & - & - & - & 21.34 & 22.79 & 25.88 & 25.70 & 26.65 & 28.50 \\
        \midrule
        \parbox[t]{6mm}{\multirow{5}{*}{\rotatebox[origin=c]{90}{\textit{150}}}} 
        &Lego    & - & - & - & - & - & – & - & - & - & 20.07 & 23.18 & 24.52 & 26.20 & 26.57 & 27.08 \\
        &Chair   & - & - & - & - & - & - & - & - & - & 20.68 & 21.47 & 31.46 & 25.77 & 26.89 & 30.75 \\
        &Ficus   & - & - & - & - & - & - & - & - & - & 22.32 & 22.01 & 23.63 & 27.20 & 29.13 & 31.98 \\
        &Hotdog  & - & - & - & - & - & - & - & - & - & 21.24 & 20.98 & 22.68 & 24.60 & 25.26 & 25.59 \\
        &Bench   & - & - & - & - & - & - & - & - & - & 20.87 & 23.28 & 24.48 & 24.03 & 24.76 & 25.57 \\
        \midrule
        &Average & - & - & - & - & - & - & - & - & - & 21.04 & 22.18 & 25.35 & 25.56 & 26.52 & 28.19 \\
        \midrule
        \parbox[t]{6mm}{\multirow{5}{*}{\rotatebox[origin=c]{90}{\textit{10}}}} 
        &Lego    & - & - & - & - & - & – & - & - & - & 16.75 & 23.32 & 15.90 & 21.85 &  23.32 & 23.72 \\
        &Chair   & - & - & - & - & - & - & - & - & - & 17.32 & 17.87 & 14.83 & 23.70 & 23.56 & 26.75 \\
        &Ficus   & - & - & - & - & - & - & - & - & - & 20.75 & 25.35 & 23.40 & 22.30 & 25.35 & 28.82 \\
        &Hotdog  & - & - & - & - & - & - & - & - & - & 16.98 & 23.81 & 15.79 & 23.03 &23.81 & 23.29 \\
        &Bench   & - & - & - & - & - & - & - & - & - & 22.90 & 23.88 & 13.75 & 22.90 & 23.87 & 24.95 \\
        \midrule
        &Average & - & - & - & - & - & - & - & - & - & 18.94 & 22.85 & 16.73 & 22.76 & 23.98 & 25.51 \\
        \bottomrule
    \end{tabular}}
\end{table*}

\begin{table*}
    \caption{Breakdown of LPIPS across all 5 simulated scenes}
    \label{tab:simulated-lpips}
    \centering
    \resizebox{\textwidth}{!}{%
    \begin{tabular}{ll|ccc|ccc|ccc|ccc|ccc}
        \toprule  
         & & \multicolumn{3}{c|}{Neuralangelo~\cite{li2023neuralangelo}} & \multicolumn{3}{c|}{RegNeRF~\cite{niemeyer2022regnerf}} & \multicolumn{3}{c|}{MonoSDF-M~\cite{yu2022monosdf}} & \multicolumn{3}{c|}{TransientNeRF~\cite{malik2023transient}} & \multicolumn{3}{c}{Ours} \\
        & \textbf{Scene} & 2 views & 3 views & 5 views & 2 views & 3 views & 5 views & 2 views & 3 views & 5 views & 2 views & 3 views & 5 views & 2 views & 3 views & 5 views \\\midrule
        \parbox[t]{6mm}{\multirow{5}{*}{\rotatebox[origin=c]{90}{\textit{6000}}}} 
        &Lego &0.480 &0.520 &0.570 &0.470 &0.460 &0.510 &0.320 &0.230 &0.190 &0.190 &0.161 &0.192 &0.197 &0.184 &0.183 \\
        &Chair &0.400 &0.530 &0.430 &0.320 &0.460 &0.410 &0.150 &0.140 &0.080 &0.138 &0.138 &0.037 &0.140 &0.136 &0.110 \\
        &Ficus &0.220 &0.230 &0.200 &0.230 &0.240 &0.220 &0.170 &0.150 &0.110 &0.094 &0.079 &0.069 &0.114 &0.106 &0.105 \\
        & Hotdog  &0.380 &0.370 &0.430 &0.440 &0.390 &0.400 &0.300 &0.330 &0.160 &0.242 &0.241 &0.118 &0.212 &0.164 &0.168 \\
        &Bench &0.460 &0.560 &0.540 &0.420 &0.510 &0.550 &0.260 &0.370 &0.120 &0.194 &0.139 &0.159 &0.216 &0.186 &0.183 \\
        \midrule
        &Average &0.388 &0.442 &0.434 &0.376 &0.412 &0.418 &0.240 &0.244 &0.132 &0.172 &0.152 &0.115 &0.176 &0.155 &0.150 \\
        \midrule
        \parbox[t]{6mm}{\multirow{5}{*}{\rotatebox[origin=c]{90}{\textit{300}}}} 
       &Lego    & - & - & - & - & - & – & - & - & - & 0.221 & 0.149 & 0.167 & 0.261 & 0.207 & 0.200 \\
        &Chair   & - & - & - & - & - & - & - & - & - & 0.156 & 0.128 & 0.040 & 0.150 & 0.130 & 0.110 \\
        &Ficus   & - & - & - & - & - & - & - & - & - & 0.119 & 0.136 & 0.138 & 0.143 & 0.127 & 0.09 \\
        &Hotdog  & - & - & - & - & - & - & - & - & - & 0.277 & 0.206 & 0.159 & 0.204 & 0.171 & 0.142 \\
        &Bench   & - & - & - & - & - & - & - & - & - & 0.232 & 0.142 & 0.100 & 0.222 & 0.177 & 0.204 \\
        \midrule
        &Average & - & - & - & - & - & - & - & - & - & 0.201 & 0.152 & 0.121 & 0.196 & 0.162 & 0.149 \\
        \midrule
        \parbox[t]{6mm}{\multirow{5}{*}{\rotatebox[origin=c]{90}{\textit{150}}}} 
        &Lego    & - & - & - & - & - & – & - & - & - & 0.220 & 0.174 & 0.160 & 0.244 & 0.194 & 0.197 \\
        &Chair   & - & - & - & - & - & - & - & - & - & 0.191 & 0.132 & 0.048 & 0.157 & 0.138 & 0.105 \\
        &Ficus   & - & - & - & - & - & - & - & - & - & 0.134 & 0.147 & 0.140 & 0.131 & 0.118 & 0.09 \\
        &Hotdog  & - & - & - & - & - & - & - & - & - & 0.256 & 0.225 & 0.127 & 0.206 & 0.182 & 0.159 \\
        &Bench   & - & - & - & - & - & - & - & - & - & 0.215 & 0.129 & 0.109 & 0.228 & 0.192 & 0.151 \\
        \midrule
        &Average & - & - & - & - & - & - & - & - & - & 0.203 & 0.161 & 0.117 & 0.193 & 0.165 & 0.151 \\
        \midrule
        \parbox[t]{6mm}{\multirow{5}{*}{\rotatebox[origin=c]{90}{\textit{10}}}} 
        &Lego    & - & - & - & - & - & – & - & - & - & 0.290 & 0.248 & 0.243 & 0.308 & 0.248 & 0.268 \\
        &Chair   & - & - & - & - & - & - & - & - & - & 0.244 & 0.185 & 0.220 & 0.210 & 0.176 & 0.152 \\
        &Ficus   & - & - & - & - & - & - & - & - & - & 0.166 & 0.185 & 0.150 & 0.221 & 0.185 & 0.124 \\
        &Hotdog  & - & - & - & - & - & - & - & - & - & 0.358 & 0.250 & 0.205 & 0.267 & 0.250 & 0.208 \\
        &Bench   & - & - & - & - & - & - & - & - & - & 0.265 & 0.220 & 0.226 & 0.265 & 0.220 & 0.289 \\
        \midrule
        &Average & - & - & - & - & - & - & - & - & - & 0.265 & 0.218 & 0.209 & 0.254 & 0.216 & 0.208 \\
        \bottomrule
    \end{tabular}}
\end{table*}

\begin{table*}
    \caption{Breakdown of SSIM across all 5 simulated scenes}
    \label{tab:simulated-ssim}
    \centering
    \resizebox{\textwidth}{!}{%
    \begin{tabular}{ll|ccc|ccc|ccc|ccc|ccc}
        \toprule  
         & & \multicolumn{3}{c|}{Neuralangelo~\cite{li2023neuralangelo}} & \multicolumn{3}{c|}{RegNeRF~\cite{niemeyer2022regnerf}} & \multicolumn{3}{c|}{MonoSDF-M~\cite{yu2022monosdf}} & \multicolumn{3}{c|}{TransientNeRF~\cite{malik2023transient}} & \multicolumn{3}{c}{Ours} \\
        & \textbf{Scene} & 2 views & 3 views & 5 views & 2 views & 3 views & 5 views & 2 views & 3 views & 5 views & 2 views & 3 views & 5 views & 2 views & 3 views & 5 views \\\midrule
        \parbox[t]{6mm}{\multirow{5}{*}{\rotatebox[origin=c]{90}{\textit{6000}}}} 
        & Lego    &0.590 &0.550 &0.480 &0.580 &0.590 &0.550 &0.730 &0.830 &0.850 &0.860 &0.897 &0.899 &0.900 &0.920 &0.930 \\
        & Chair   &0.640 &0.530 &0.680 &0.730 &0.580 &0.680 &0.870 &0.870 &0.950 &0.901 &0.899 &0.977 &0.930 &0.940 &0.960 \\
        & Ficus   &0.820 &0.810 &0.860 &0.800 &0.800 &0.830 &0.870 &0.880 &0.930 &0.925 &0.934 &0.942 &0.950 &0.942 &0.969 \\
        & Hotdog  &0.640 &0.670 &0.650 &0.570 &0.630 &0.660 &0.810 &0.780 &0.870 &0.882 &0.875 &0.963 &0.900 &0.942 &0.960 \\
        & Bench   &0.580 &0.450 &0.530 &0.600 &0.540 &0.450 &0.770 &0.720 &0.870 &0.856 &0.884 &0.870 &0.870 &0.873 &0.890 \\
        \midrule
        &Average &0.654 &0.602 &0.640 &0.656 &0.628 &0.634 &0.810 &0.816 &0.894 &0.885 &0.898 &0.930 &0.910 &0.923 &0.942\\
        \midrule
        \parbox[t]{6mm}{\multirow{5}{*}{\rotatebox[origin=c]{90}{\textit{300}}}} 
        &Lego   & - & - & - & - & - & – & - & - & - & 0.846 & 0.903 & 0.910 & 0.893 & 0.915 & 0.919 \\
        &Chair   & - & - & - & - & - & - & - & - & - & 0.891 & 0.892 & 0.974 & 0.925 & 0.936 & 0.961 \\
        &Ficus   & - & - & - & - & - & - & - & - & - & 0.895 & 0.879 & 0.883 & 0.935 & 0.943 & 0.962 \\
        &Hotdog  & - & - & - & - & - & - & - & - & - & 0.859 & 0.901 & 0.908 & 0.925 & 0.939 & 0.954 \\
        &Bench   & - & - & - & - & - & - & - & - & - & 0.842 & 0.886 & 0.896 & 0.861 & 0.872 & 0.878 \\
        \midrule
        &Average & - & - & - & - & - & - & - & - & - & 0.867 & 0.892 & 0.914 & 0.907 & 0.921 & 0.935 \\
        \midrule
        \parbox[t]{6mm}{\multirow{5}{*}{\rotatebox[origin=c]{90}{\textit{150}}}} 
        &Lego    & - & - & - & - & - & – & - & - & - & 0.840 & 0.882 & 0.896 & 0.895 & 0.913 & 0.920 \\
        &Chair   & - & - & - & - & - & - & - & - & - & 0.863 & 0.881 & 0.964 & 0.923 & 0.934 & 0.960 \\
        &Ficus   & - & - & - & - & - & - & - & - & - & 0.882 & 0.867 & 0.879 & 0.930 & 0.947 & 0.965 \\
        &Hotdog  & - & - & - & - & - & - & - & - & - & 0.874 & 0.855 & 0.911 & 0.930 & 0.934 & 0.946 \\
        &Bench   & - & - & - & - & - & - & - & - & - & 0.851 & 0.889 & 0.893 & 0.857 & 0.871 & 0.873 \\
        \midrule
        &Average & - & - & - & - & - & - & - & - & - & 0.862 & 0.875 & 0.909 & 0.907 & 0.920 & 0.933 \\
        \midrule
        \parbox[t]{6mm}{\multirow{5}{*}{\rotatebox[origin=c]{90}{\textit{10}}}} 
        &Lego    & - & - & - & - & - & – & - & - & - & 0.745 & 0.840 & 0.737 & 0.797 & 0.840 & 0.840 \\
        &Chair   & - & - & - & - & - & - & - & - & - & 0.801 & 0.816 & 0.778 & 0.872 & 0.876 & 0.876 \\
        &Ficus   & - & - & - & - & - & - & - & - & - & 0.850 & 0.877 & 0.866 & 0.871 & 0.877 & 0.877 \\
        &Hotdog  & - & - & - & - & - & - & - & - & - & 0.739 & 0.860 & 0.789  & 0.852 & 0.860 & 0.860 \\
        &Bench   & - & - & - & - & - & - & - & - & - & 0.823 & 0.844 & 0.762 & 0.822 & 0.844 & 0.844 \\
        \midrule
        &Average & - & - & - & - & - & - & - & - & - & 0.792 & 0.847 & 0.786 & 0.843 & 0.859 & 0.859 \\
        \bottomrule
    \end{tabular}}
\end{table*}

\begin{table*}
    \caption{Breakdown of PSNR across all 6 captured scenes}
    \label{tab:captured-psnr}
    \centering
    \resizebox{\textwidth}{!}{%
    \begin{tabular}{ll|ccc|ccc|ccc|ccc|ccc}
        \toprule  
         & & \multicolumn{3}{c|}{Neuralangelo~\cite{li2023neuralangelo}} & \multicolumn{3}{c|}{RegNeRF~\cite{niemeyer2022regnerf}} & \multicolumn{3}{c|}{MonoSDF-M~\cite{yu2022monosdf}} & \multicolumn{3}{c|}{TransientNeRF~\cite{malik2023transient}} & \multicolumn{3}{c}{Ours} \\
        & \textbf{Scene} & 2 views & 3 views & 5 views & 2 views & 3 views & 5 views & 2 views & 3 views & 5 views & 2 views & 3 views & 5 views & 2 views & 3 views & 5 views \\\midrule
        \parbox[t]{6mm}{\multirow{6}{*}{\rotatebox[origin=c]{90}{\textit{1500}}}} 
        & Cinema  &17.80 &18.46 &20.39 &17.96 &19.63 &17.03 &18.53 &21.71 &25.75 &21.61 &21.66 &25.12 &23.63 &24.70 &28.03 \\
        & Food    &19.09 &19.51 &23.03 &19.89 &20.44 &22.42 &17.31 &22.34 &30.74 &23.40 &23.78 &22.09 &24.58 &24.53 &25.12 \\
        & Carving &19.36 &21.44 &22.17 &19.63 &21.28 &21.47 &19.27 &21.85 &27.00 &23.52 &24.20 &24.70 &20.62 &25.65 &27.92 \\
        & Boots   &15.89 &17.86 &19.93 &17.38 &17.74 &18.92 &15.77 &22.59 &30.30 &22.38 &22.29 &24.94 &16.58 &25.43 &26.32 \\
        & Baskets &19.92 &23.07 &24.14 &20.95 &23.28 &24.86 &20.43 &21.12 &24.31 &22.48 &20.93 &19.90 &22.43 &23.10 &23.32 \\
        & Chef    &14.69 &15.32 &16.76 &15.20 &15.15 &17.03 &14.47 &17.35 &25.38 &19.27 &18.14 &19.55 &20.02 &20.30 &20.61 \\
        \midrule
        &Average &17.79 &19.28 &21.07 &18.50 &19.59 &20.29 &17.63 &21.16 &27.25 &22.11 &21.83 &22.72 &21.31 &23.95 &25.22 \\
        \midrule
        
        \parbox[t]{6mm}{\multirow{5}{*}{\rotatebox[origin=c]{90}{\textit{300}}}} 
        & Cinema  & - & - & - & - & - & - & - & - & - & 16.82 &22.37 &22.43 &21.94 &23.76 &28.08 \\
        & Food   & - & - & - & - & - & - & - & - & - &18.86 &22.31 &21.09 &22.98 &23.00 &24.03 \\
        & Carving & - & - & - & - & - & - & - & - & -&23.48 &23.83 &25.39 &21.73 &25.48 &27.82 \\
        & Boots  & - & - & - & - & - & - & - & - & - &24.37 &22.53 &18.07 &20.86 &25.32 &17.57 \\
        & Baskets & - & - & - & - & - & - & - & - & -&22.22 &22.15 &20.27 &21.32 &22.97 &24.24 \\
        & Chef   & - & - & - & - & - & - & - & - & - &15.09 &18.72 &20.53 &20.78 &20.66 &20.98 \\
        \midrule
        &Average & - & - & - & - & - & - & - & - & -&20.14 &21.98 &21.30 &21.60 &23.53 &23.79 \\
        \midrule
        \parbox[t]{6mm}{\multirow{5}{*}{\rotatebox[origin=c]{90}{\textit{150}}}} 
        & Cinema  & - & - & - & - & - & - & - & - & - &17.06 &21.97 &19.18 &22.44 &24.58 &27.86 \\
        & Food    & - & - & - & - & - & - & - & - & -&18.80 &20.47 &21.98 &22.93 &23.05 &24.18 \\
        & Carving & - & - & - & - & - & - & - & - & -&18.36 &24.14 &25.19 &18.96 &25.39 &27.75 \\
        & Boots   & - & - & - & - & - & - & - & - & -&23.32 &24.36 &19.24 &20.92 &24.91 &26.91 \\
        & Baskets & - & - & - & - & - & - & - & - & -&22.14 &19.41 &18.28 &21.53 &23.15 &24.14 \\
        & Chef    & - & - & - & - & - & - & - & - & -&15.07 &16.02 &15.80 &20.61 &20.28 &21.13 \\
        \midrule
        & Average & - & - & - & - & - & - & - & - & - &19.12 &21.06 &19.94 &21.23 &23.56 &25.33 \\
        \midrule
        \parbox[t]{6mm}{\multirow{5}{*}{\rotatebox[origin=c]{90}{\textit{10}}}} 
        & Cinema & - & - & - & - & - & - & - & - & - &15.36 &15.21 &14.33 &21.91 &23.06 &25.30 \\
        & Food  & - & - & - & - & - & - & - & - & -  &18.44 &18.27 &17.39 &22.71 &23.05 &24.43 \\
        & Carving & - & - & - & - & - & - & - & - & -&17.58 &19.59 &17.71 &23.07 &24.76 &27.37 \\
        & Boots  & - & - & - & - & - & - & - & - & - &16.32 &15.09 &14.81 &23.83 &24.29 &28.35 \\
        & Baskets & - & - & - & - & - & - & - & - & -&16.58 &16.38 &15.38 &23.10 &22.85 &24.63 \\
        & Chef  & - & - & - & - & - & - & - & - & -  &14.13 &13.63 &12.79 &19.79 &20.26 &22.76 \\
         \midrule
        &Average &- & - & - & - & - & - & - & - & -&16.40 &16.36 &15.40 &22.40 &23.04 &25.47\\
        \bottomrule
    \end{tabular}}
\end{table*}

\begin{table*}
    \caption{Breakdown of LPIPS across all 6 captured scenes}
    \label{tab:captured-lpips}
    \centering
    \resizebox{\textwidth}{!}{%
    \begin{tabular}{ll|ccc|ccc|ccc|ccc|ccc}
        \toprule  
         & & \multicolumn{3}{c|}{Neuralangelo~\cite{li2023neuralangelo}} & \multicolumn{3}{c|}{RegNeRF~\cite{niemeyer2022regnerf}} & \multicolumn{3}{c|}{MonoSDF-M~\cite{yu2022monosdf}} & \multicolumn{3}{c|}{TransientNeRF~\cite{malik2023transient}} & \multicolumn{3}{c}{Ours} \\
        & \textbf{Scene} & 2 views & 3 views & 5 views & 2 views & 3 views & 5 views & 2 views & 3 views & 5 views & 2 views & 3 views & 5 views & 2 views & 3 views & 5 views \\\midrule
        \parbox[t]{6mm}{\multirow{5}{*}{\rotatebox[origin=c]{90}{\textit{1500}}}} 
        & Cinema  &0.424 &0.387 &0.350 &0.390 &0.330 &0.470 &0.330 &0.260 &0.180 &0.281 &0.245 &0.178 &0.223 &0.163 &0.138 \\
        & Food    &0.384 &0.368 &0.247 &0.400 &0.330 &0.270 &0.370 &0.230 &0.130 &0.286 &0.205 &0.154 &0.183 &0.176 &0.141 \\
        & Carving &0.266 &0.244 &0.205 &0.260 &0.220 &0.220 &0.270 &0.230 &0.130 &0.232 &0.138 &0.103 &0.222 &0.115 &0.100 \\
        & Boots   &0.289 &0.369 &0.261 &0.300 &0.320 &0.290 &0.380 &0.160 &0.110 &0.221 &0.182 &0.155 &0.302 &0.128 &0.111 \\
        & Baskets &0.318 &0.298 &0.268 &0.320 &0.270 &0.250 &0.320 &0.220 &0.160 &0.269 &0.165 &0.164 &0.211 &0.179 &0.176 \\
        & Chef    &0.601 &0.464 &0.467 &0.580 &0.450 &0.460 &0.450 &0.390 &0.230 &0.334 &0.338 &0.276 &0.302 &0.257 &0.226 \\
        \midrule
        &Average &0.380 &0.355 &0.300 &0.375 &0.320 &0.327 &0.353 &0.248 &0.157 &0.270 &0.212 &0.172 &0.240 &0.170 &0.149 \\
        \midrule
        
        \parbox[t]{6mm}{\multirow{5}{*}{\rotatebox[origin=c]{90}{\textit{300}}}} 
        & Cinema  & - & - & - & - & - & - & - & - & - &0.356 &0.204 &0.189 &0.238 &0.182 &0.148 \\
        & Food    & - & - & - & - & - & - & - & - & - &0.256 &0.191 &0.186 &0.180 &0.169 &0.140 \\
        & Carving & - & - & - & - & - & - & - & - & - &0.209 &0.166 &0.104 &0.225 &0.127 &0.088 \\
        & Boots   & - & - & - & - & - & - & - & - & - &0.223 &0.200 &0.171 &0.241 &0.135 &0.153 \\
        & Baskets & - & - & - & - & - & - & - & - & - &0.222 &0.183 &0.177 &0.184 &0.163 &0.163 \\
        & Chef   & - & - & - & - & - & - & - & - & -  &0.356 &0.315 &0.234 &0.269 &0.244 &0.214 \\
        \midrule
        &Average & - & - & - & - & - & - & - & - & - &0.270 &0.210 &0.177 &0.223 &0.170 &0.151 \\
        \midrule
        \parbox[t]{6mm}{\multirow{5}{*}{\rotatebox[origin=c]{90}{\textit{150}}}} 
        & Cinema  & - & - & - & - & - & - & - & - & - &0.356 &0.232 &0.219 &0.212 &0.183 &0.144 \\
        & Food    & - & - & - & - & - & - & - & - & - &0.236 &0.171 &0.170 &0.188 &0.178 &0.134 \\
        & Carving & - & - & - & - & - & - & - & - & - &0.224 &0.181 &0.106 &0.263 &0.114 &0.094 \\
        & Boots   & - & - & - & - & - & - & - & - & - &0.247 &0.113 &0.168 &0.216 &0.123 &0.117 \\
        & Baskets & - & - & - & - & - & - & - & - & - &0.261 &0.210 &0.193 &0.192 &0.182 &0.166 \\
        & Chef    & - & - & - & - & - & - & - & - & - &0.337 &0.326 &0.317 &0.260 &0.245 &0.220 \\
        \midrule
        & Average & - & - & - & - & - & - & - & - & - &0.277 &0.206 &0.196 &0.222 &0.171 &0.146 \\
        \midrule
        \parbox[t]{6mm}{\multirow{5}{*}{\rotatebox[origin=c]{90}{\textit{10}}}} 
        & Cinema & - & - & - & - & - & - & - & - & - &0.266 &0.243 &0.333 &0.305 &0.238 &0.193 \\
        & Food   & - & - & - & - & - & - & - & - & - &0.225 &0.218 &0.227 &0.205 &0.202 &0.162 \\
        & Carving & - & - & - & - & - & - & - & - & -&0.200 &0.150 &0.148 &0.193 &0.139 &0.114 \\
        & Boots  & - & - & - & - & - & - & - & - & - &0.213 &0.396 &0.270 &0.183 &0.150 &0.124 \\
        & Baskets & - & - & - & - & - & - & - & - & -&0.279 &0.237 &0.290 &0.243 &0.217 &0.197 \\
        & Chef  & - & - & - & - & - & - & - & - & -  &0.290 &0.307 &0.361 &0.321 &0.300 &0.250 \\
        \midrule
        &Average & - & - & - & - & - & - & - & - & -&0.246 &0.258 &0.272 &0.242 &0.208 &0.173\\
        \bottomrule
    \end{tabular}}
\end{table*}

\begin{table*}
    \caption{Breakdown of SSIM across all 6 captured scenes}
    \label{tab:captured-ssim}
    \centering
    \resizebox{\textwidth}{!}{%
    \begin{tabular}{ll|ccc|ccc|ccc|ccc|ccc}
        \toprule  
         & & \multicolumn{3}{c|}{Neuralangelo~\cite{li2023neuralangelo}} & \multicolumn{3}{c|}{RegNeRF~\cite{niemeyer2022regnerf}} & \multicolumn{3}{c|}{MonoSDF-M~\cite{yu2022monosdf}} & \multicolumn{3}{c|}{TransientNeRF~\cite{malik2023transient}} & \multicolumn{3}{c}{Ours} \\
        & \textbf{Scene} & 2 views & 3 views & 5 views & 2 views & 3 views & 5 views & 2 views & 3 views & 5 views & 2 views & 3 views & 5 views & 2 views & 3 views & 5 views \\\midrule
        \parbox[t]{6mm}{\multirow{5}{*}{\rotatebox[origin=c]{90}{\textit{1500}}}} 
        & Cinema  &0.600 &0.690 &0.710 &0.600 &0.650 &0.480 &0.730 &0.810 &0.910 &0.850 &0.812 &0.879 &0.879 &0.897 &0.925 \\
        & Food    &0.650 &0.760 &0.840 &0.500 &0.610 &0.720 &0.610 &0.830 &0.950 &0.826 &0.879 &0.873 &0.890 &0.892 &0.912 \\
        & Carving &0.740 &0.820 &0.850 &0.700 &0.760 &0.770 &0.770 &0.840 &0.930 &0.913 &0.929 &0.929 &0.804 &0.933 &0.951 \\
        & Boots   &0.740 &0.670 &0.800 &0.670 &0.640 &0.690 &0.640 &0.900 &0.960 &0.912 &0.909 &0.914 &0.706 &0.933 &0.940 \\
        & Baskets &0.700 &0.760 &0.790 &0.630 &0.700 &0.750 &0.690 &0.810 &0.900 &0.846 &0.850 &0.826 &0.854 &0.860 &0.870 \\
        & Chef    &0.370 &0.600 &0.600 &0.290 &0.500 &0.480 &0.600 &0.670 &0.910 &0.775 &0.747 &0.811 &0.823 &0.847 &0.856 \\
        \midrule
        &Average &0.633 &0.717 &0.765 &0.565 &0.643 &0.648 &0.673 &0.810 &0.927 &0.854 &0.854 &0.872 &0.826 &0.894 &0.909 \\
        \midrule

        \parbox[t]{6mm}{\multirow{5}{*}{\rotatebox[origin=c]{90}{\textit{300}}}} 
        & Cinema  & - & - & - & - & - & - & - & - & -&0.585 &0.851 &0.859 &0.854 &0.882 &0.922 \\
        & Food    & - & - & - & - & - & - & - & - & -&0.813 &0.877 &0.840 &0.884 &0.889 &0.908 \\
        & Carving & - & - & - & - & - & - & - & - & -&0.911 &0.913 &0.935 &0.846 &0.930 &0.952 \\
        & Boots   & - & - & - & - & - & - & - & - & -&0.923 &0.899 &0.851 &0.809 &0.932 &0.869 \\
        & Baskets & - & - & - & - & - & - & - & - & -&0.859 &0.860 &0.824 &0.845 &0.867 &0.885 \\
        & Chef    & - & - & - & - & - & - & - & - & -&0.717 &0.751 &0.849 &0.846 &0.854 &0.867 \\
        \midrule
        &Average & - & - & - & - & - & - & - & - & -&0.801 &0.858 &0.860 &0.847 &0.892 &0.900 \\
        \midrule
        \parbox[t]{6mm}{\multirow{5}{*}{\rotatebox[origin=c]{90}{\textit{150}}}} 
        & Cinema  & - & - & - & - & - & - & - & - & -&0.663 &0.842 &0.810 &0.863 &0.889 &0.920 \\
        & Food   & - & - & - & - & - & - & - & - & - &0.816 &0.855 &0.878 &0.882 &0.884 &0.912 \\
        & Carving & - & - & - & - & - & - & - & - & -&0.825 &0.912 &0.933 &0.762 &0.931 &0.950 \\
        & Boots  & - & - & - & - & - & - & - & - & - &0.899 &0.930 &0.878 &0.834 &0.932 &0.942 \\
        & Baskets & - & - & - & - & - & - & - & - & -&0.851 &0.804 &0.791 &0.851 &0.865 &0.886 \\
        & Chef   & - & - & - & - & - & - & - & - & - &0.704 &0.716 &0.699 &0.844 &0.848 &0.862 \\
        \midrule
        &Average & - & - & - & - & - & - & - & - & -&0.793 &0.843 &0.832 &0.839 &0.892 &0.912 \\
        \midrule
        \parbox[t]{6mm}{\multirow{5}{*}{\rotatebox[origin=c]{90}{\textit{10}}}} 
        & Cinema & - & - & - & - & - & - & - & - & - &0.744 &0.741 &0.457 &0.802 &0.841 &0.883 \\
        & Food & - & - & - & - & - & - & - & - & -   &0.808 &0.789 &0.652 &0.868 &0.871 &0.907 \\
        & Carving & - & - & - & - & - & - & - & - & -&0.850 &0.872 &0.854 &0.881 &0.918 &0.938 \\
        & Boots  & - & - & - & - & - & - & - & - & - &0.833 &0.163 &0.519 &0.897 &0.905 &0.929 \\
        & Baskets & - & - & - & - & - & - & - & - & -&0.765 &0.761 &0.517 &0.845 &0.844 &0.874 \\
        & Chef  & - & - & - & - & - & - & - & - & -  &0.714 &0.698 &0.451 &0.792 &0.806 &0.847 \\
        \midrule
        &Average & - & - & - & - & - & - & - & - & -&0.786 &0.671 &0.575 &0.848 &0.864 &0.896 \\
        \bottomrule
    \end{tabular}}
\end{table*}

\begin{table*}
    \caption{Breakdown of L1 Depth across all 6 captured scenes}
    \label{tab:captured-l1depth}
    \centering
    \resizebox{\textwidth}{!}{%
    \begin{tabular}{ll|ccc|ccc|ccc|ccc|ccc}
        \toprule  
         & & \multicolumn{3}{c|}{Neuralangelo~\cite{li2023neuralangelo}} & \multicolumn{3}{c|}{RegNeRF~\cite{niemeyer2022regnerf}} & \multicolumn{3}{c|}{MonoSDF-M~\cite{yu2022monosdf}} & \multicolumn{3}{c|}{TransientNeRF~\cite{malik2023transient}} & \multicolumn{3}{c}{Ours} \\
        & \textbf{Scene} & 2 views & 3 views & 5 views & 2 views & 3 views & 5 views & 2 views & 3 views & 5 views & 2 views & 3 views & 5 views & 2 views & 3 views & 5 views \\\midrule
        \parbox[t]{6mm}{\multirow{5}{*}{\rotatebox[origin=c]{90}{\textit{1500}}}} 
        & Cinema  &0.080 &0.068 &0.068 &0.080 &0.070 &0.100 &0.030 &0.030 &0.030 &0.006 &0.006 &0.007 &0.007 &0.007 &0.007 \\
        & Food    &0.094 &0.080 &0.069 &0.090 &0.080 &0.090 &0.040 &0.020 &0.030 &0.006 &0.007 &0.016 &0.006 &0.007 &0.008 \\
        & Carving &0.054 &0.060 &0.045 &0.070 &0.060 &0.060 &0.030 &0.030 &0.020 &0.006 &0.005 &0.006 &0.007 &0.006 &0.006 \\
        & Boots   &0.069 &0.051 &0.045 &0.060 &0.050 &0.040 &0.020 &0.005 &0.004 &0.001 &0.002 &0.002 &0.010 &0.001 &0.001 \\
        & Baskets &0.074 &0.069 &0.049 &0.090 &0.080 &0.070 &0.050 &0.040 &0.030 &0.008 &0.008 &0.026 &0.007 &0.008 &0.011 \\
        & Chef    &0.086 &0.100 &0.090 &0.120 &0.100 &0.100 &0.030 &0.030 &0.010 &0.003 &0.006 &0.006 &0.002 &0.004 &0.005 \\
        \midrule
        &Average &0.076 &0.071 &0.061 &0.085 &0.073 &0.077 &0.033 &0.026 &0.021 &0.005 &0.006 &0.010 &0.006 &0.006 &0.006 \\
        \midrule

        \parbox[t]{6mm}{\multirow{5}{*}{\rotatebox[origin=c]{90}{\textit{300}}}} 
        & Cinema  & - & - & - & - & - & - & - & - & - &0.028 &0.010 &0.018 &0.007 &0.007 &0.007 \\
        & Food   &- & - & - & - & - & - & - & - & -   &0.020 &0.010 &0.015 &0.006 &0.006 &0.007 \\
        & Carving & - & - & - & - & - & - & - & - & - &0.007 &0.007 &0.007 &0.007 &0.006 &0.006 \\
        & Boots  &- & - & - & - & - & - & - & - & -   &0.001 &0.002 &0.009 &0.001 &0.001 &0.034 \\
        & Baskets & - & - & - & - & - & - & - & - & - &0.008 &0.009 &0.029 &0.007 &0.007 &0.011 \\
        & Chef   & - & - & - & - & - & - & - & - & -  &0.010 &0.005 &0.006 &0.002 &0.004 &0.004 \\
        \midrule
        &Average & - & - & - & - & - & - & - & - & - &0.012 &0.007 &0.014 &0.005 &0.005 &0.012 \\
        \midrule
        \parbox[t]{6mm}{\multirow{5}{*}{\rotatebox[origin=c]{90}{\textit{150}}}} 
        & Cinema & - & - & - & - & - & - & - & - & - &0.028 &0.011 &0.021 &0.007 &0.007 &0.007 \\
        & Food  & - & - & - & - & - & - & - & - & -  &0.023 &0.014 &0.014 &0.006 &0.006 &0.007 \\
        & Carving & - & - & - & - & - & - & - & - & -&0.024 &0.007 &0.006 &0.014 &0.006 &0.006 \\
        & Boots  & - & - & - & - & - & - & - & - & - &0.001 &0.001 &0.008 &0.004 &0.001 &0.001 \\
        & Baskets& - & - & - & - & - & - & - & - & - &0.009 &0.033 &0.035 &0.007 &0.007 &0.009 \\
        & Chef  & - & - & - & - & - & - & - & - & -  &0.012 &0.022 &0.022 &0.001 &0.005 &0.004 \\
        \midrule
        &Average & - & - & - & - & - & - & - & - & -&0.016 &0.015 &0.018 &0.006 &0.005 &0.006 \\
        \midrule
        \parbox[t]{6mm}{\multirow{5}{*}{\rotatebox[origin=c]{90}{\textit{10}}}} 
        & Cinema & - & - & - & - & - & - & - & - & - &0.045 &0.040 &0.117 &0.009 &0.011 &0.008 \\
        & Food  & - & - & - & - & - & - & - & - & -  &0.033 &0.034 &0.086 &0.008 &0.008 &0.008 \\
        & Carving & - & - & - & - & - & - & - & - & -&0.028 &0.014 &0.023 &0.010 &0.008 &0.007 \\
        & Boots & - & - & - & - & - & - & - & - & -  &0.017 &0.067 &0.073 &0.001 &0.001 &0.001 \\
        & Baskets & - & - & - & - & - & - & - & - & -&0.049 &0.054 &0.112 &0.016 &0.015 &0.014 \\
        & Chef  & - & - & - & - & - & - & - & - & -  &0.049 &0.037 &0.131 &0.006 &0.005 &0.004 \\
        \midrule
        &Average & - & - & - & - & - & - & - & - & -&0.037 &0.041 &0.090 &0.008 &0.008 &0.00 \\
        \bottomrule
    \end{tabular}}
\end{table*}

\begin{figure*}[t]
    \centering
    \includegraphics[width=0.75\textwidth]{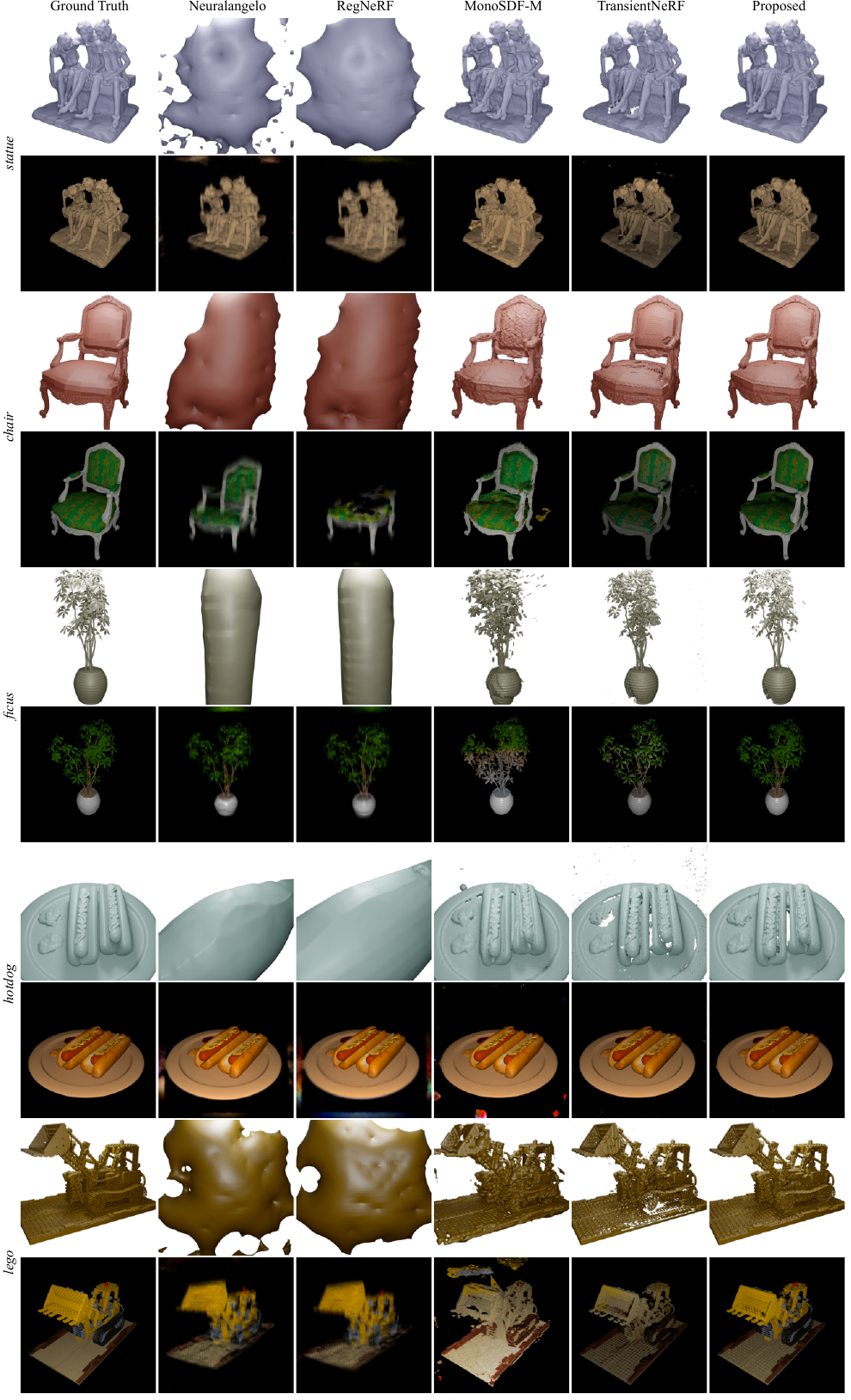}
    \caption{Rendered meshes and images on the simulated dataset for 2 views.}
    \label{fig:sim2}
\end{figure*}

\begin{figure*}[t]
    \centering
    \includegraphics[width=0.75\textwidth]{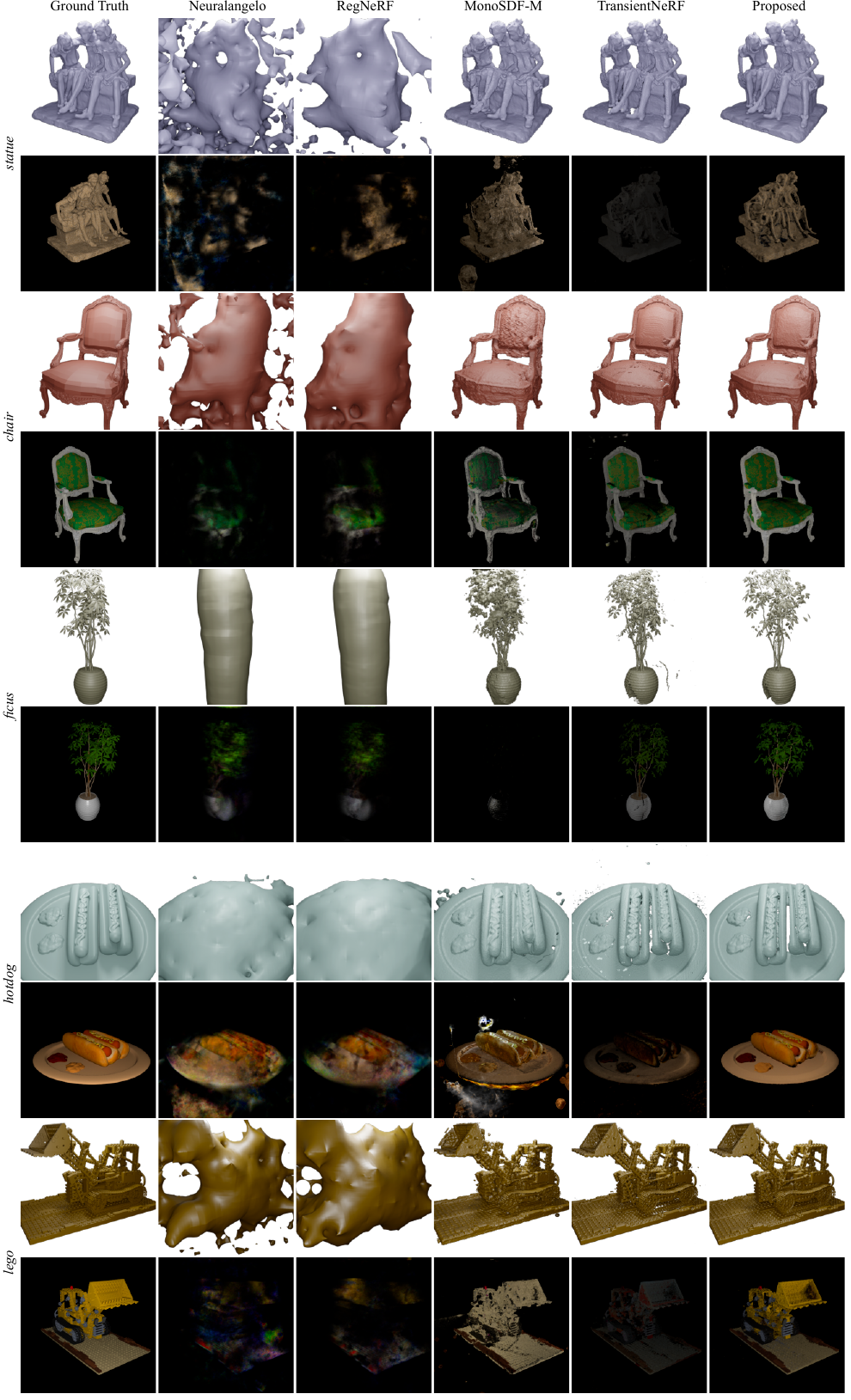}
    \caption{Rendered meshes and images on the simulated dataset for 3 views.}
    \label{fig:sim3}
\end{figure*}

\begin{figure*}[t]
    \centering
    \includegraphics[width=0.75\textwidth]{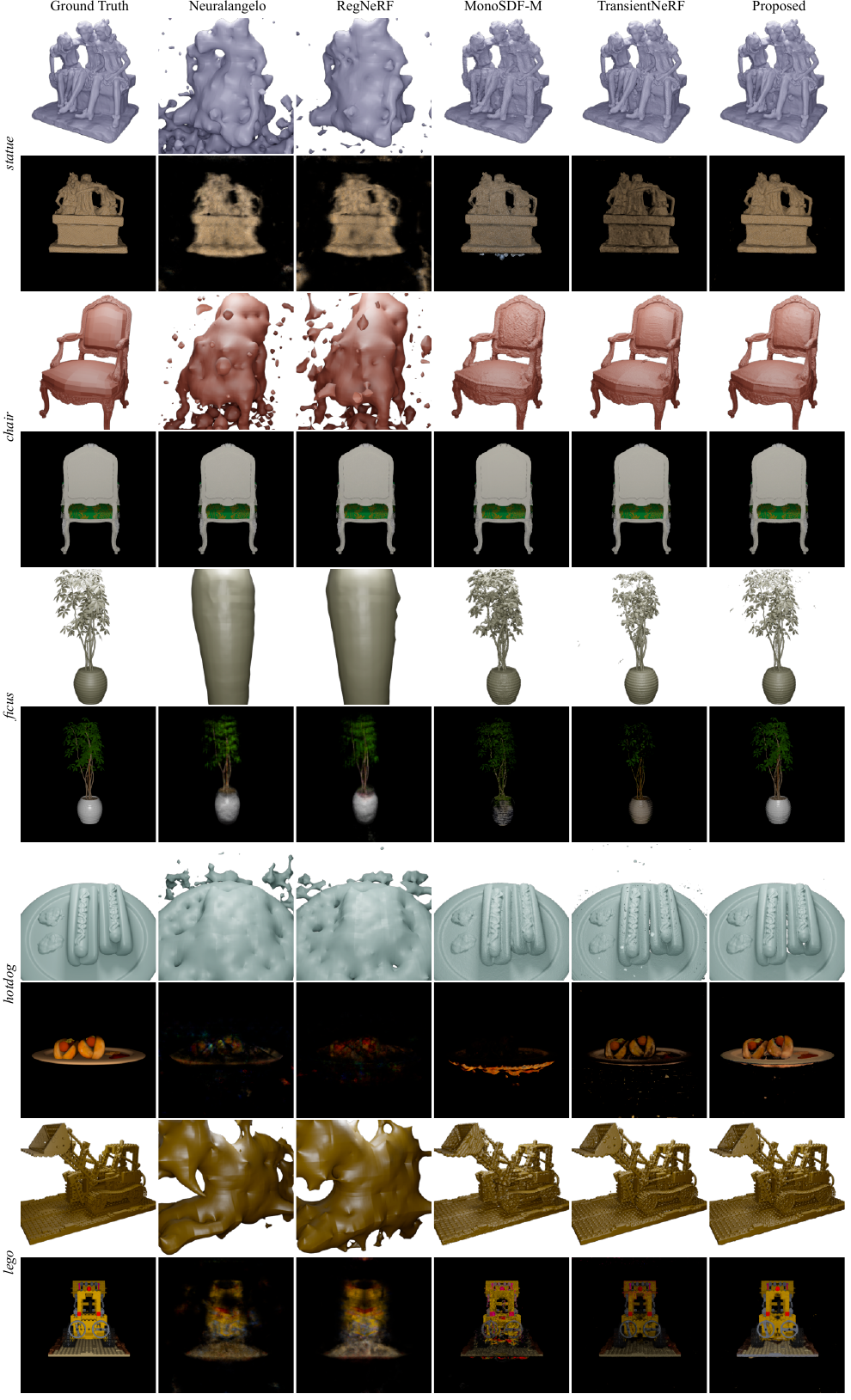}
    \caption{Rendered meshes and images on the simulated dataset for 5 views.}
    \label{fig:sim5}
\end{figure*}

\end{document}